\documentclass[twoside,11pt]{article}

\usepackage{blindtext}

\usepackage[abbrvbib, preprint]{jmlr2e}

\usepackage{amsmath}
\usepackage{algorithm}  
\usepackage{algorithmic} 
\usepackage{multirow}
\usepackage{dsfont}
\usepackage{threeparttable}

\newcommand*{\Scale}[2][4]{\scalebox{#1}{$#2$}}

\usepackage{color}
\definecolor{Red}{rgb}{1,0,0}

\usepackage{lastpage}
\jmlrheading{xx}{2022}{1-\pageref{LastPage}}{x/xx; Revised x/xx}{x/xx}{21-0000}{A. Wu, K. Kuang, R. Xiong, B. Li and F. Wu}

\ShortHeadings{Confounder Balancing for Instrumental Variable Regression with Latent Variable}{Wu, Kuang, Xiong, Li and Wu}
\firstpageno{1}

\begin{document}

\title{Confounder Balancing for Instrumental Variable Regression with Latent Variable}

\author{\name Anpeng Wu \email anpwu@zju.edu.cn \\
        \addr Department of Computer Science and Technology, Zhejiang University, Hangzhou, 310007 China
        \AND
        \name Kun Kuang \email kunkuang@zju.edu.cn \\
        \addr Department of Computer Science and Technology, Zhejiang University, Hangzhou, 310007 China
        \AND
        \name Ruoxuan Xiong \email ruoxuan.xiong@emory.edu \\
        \addr Department of Quantitative Theory and Methods, Emory University, Atlanta, 30322 U.S.
        \AND
        \name Bo Li \email libo@sem.tsinghua.edu.cn \\
        \addr School of Economics and Management, Tsinghua University, Beijing, 100084 China
        \AND
        \name Fei Wu \email wufei@zju.edu.cn \\
        \addr Department of Computer Science and Technology, Zhejiang University, Hangzhou, 310007 China
       }

\editor{My editor}

\maketitle

\renewcommand{\thefootnote}{\fnsymbol{footnote}}
\renewcommand{\thefootnote}{\arabic{footnote}}

\begin{abstract}
This paper studies the confounding effects from the unmeasured confounders and the imbalance of observed confounders in IV regression and aims at unbiased causal effect estimation. Recently, nonlinear IV estimators were proposed to allow for nonlinear model in both stages. However, the observed confounders may be imbalanced in stage 2, which could still lead to biased treatment effect estimation in certain cases. To this end, we propose a Confounder Balanced IV Regression (CB-IV) algorithm to jointly remove the bias from the unmeasured confounders and the imbalance of observed confounders. Theoretically, by redefining and solving an inverse problem for potential outcome function, we show that our CB-IV algorithm can unbiasedly estimate treatment effects and achieve lower variance.

The IV methods have a major disadvantage in that little prior or theory is currently available to pre-define a valid IV in real-world scenarios. Thus, we study two more challenging settings without pre-defined valid IVs: (1) indistinguishable IVs implicitly present in observations, i.e., mixed-variable challenge, and (2) latent IVs don’t appear in observations, i.e., latent-variable challenge. To address these two challenges, we extend our CB-IV by a latent-variable module, namely CB-IV-L algorithm. Extensive experiments demonstrate that our CB-IV(-L) outperforms the existing approaches.
\end{abstract}


\section{Introduction}

Treatment effect estimation is a fundamental problem in causal inference, and its key challenge is to remove the confounding bias induced by the confounders, which affect both treatment and outcome\footnote{As introduced in Chapter 3.3 in \citet{pearl2009causality}, the \textbf{confounding bias} between the treatment and outcome can be defined as the bias of treatment effect estimation when imbalanced confounders exist. }. 
Under the unconfoundedness assumption (i.e., no unmeasured confounders), many confounder balancing methods, such as \citet{35rubin1973matching,32li2016matching,45shalit2017estimating,jung2021estimating}, have been proposed to break the dependence between the treatment and all confounders. 
In practice, however, the unconfoundedness assumption is hardly satisfied, and unmeasured confounders always exist \citep{spirtes2010introduction,salehkaleybar2020learning}. How to precisely estimate the treatment effect from observational data in the presence of unmeasured confounders is of vital importance for both academic research and real applications.

\begin{figure}
\begin{center}
\includegraphics[width=0.95\linewidth]{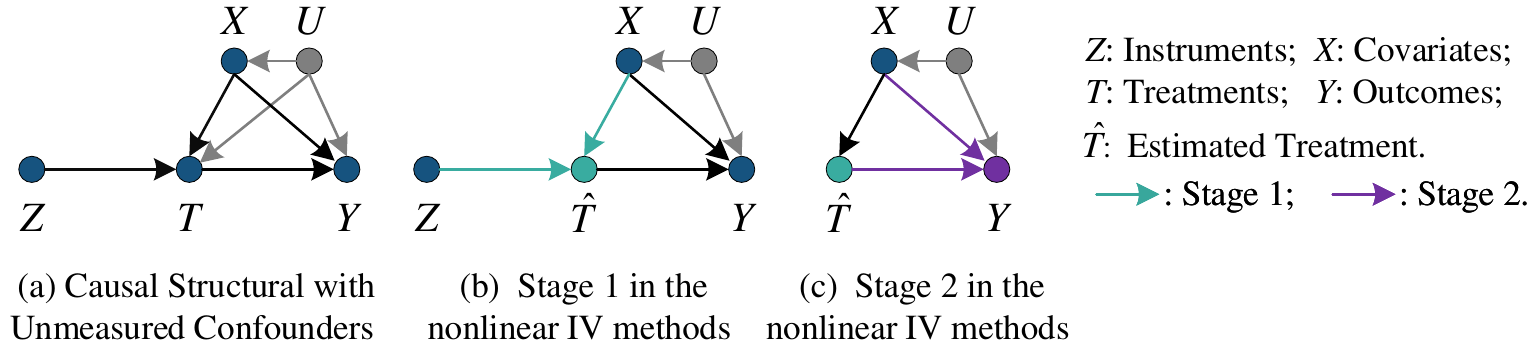}
\end{center}
\vspace{-12pt}
\caption{(a) Causal structural with unmeasured confounders; (b) Treatment regression stage in the nonlinear IV methods; (c) Outcome regression stage in the nonlinear IV methods: conventional IV methods cut off the effect of $U$ on $\hat{T}$, however, the observed confounders would affect both the estimated treatment $\hat{T}$ and outcome $Y$, leading to confounding bias in outcome regression {when the outcome model is misspecified}. Blue nodes denote observable variables, and gray nodes indicate unmeasured variables. The gray lines encode the effects from unmeasured variables.}
\label{figure1}
\vspace{-15pt}
\end{figure}

A classical method to address the bias induced by unmeasured confounders is IV regression methods \citep{14heckman2008econometric,jung2021double,gupta2021efficient}. 
As shown in Figure~\ref{figure1}(a), let $T$ be the treatment, $Y$ be the outcome of interest, $X$ and $U$ be the observed and unobserved confounders, respectively, where $U$ might affect or be affected by $X$. $Z$ denotes the instrumental variables (IVs), which only influence $Y$ via $T$. In the presence of unmeasured confounders, two-stage least squares (2SLS) regression and its variants \citep{16angrist1995identification,17angrist2001instrumental,26buhlmann2014cam} can identify the treatment effect with the following two stages. 
{In stage 1, they regress the treatment $T$ on instruments $Z$; in stage 2, they regress the outcome $Y$ on the fitted $\hat{T}$ from stage 1. }
However, these methods are limited to the linear models and require additive noise assumption, i.e., the effect of unmeasured confounders on outcome is an additive zero-mean noise \citep{angrist1996identification,19stock2002survey,26buhlmann2014cam}.

For nonlinear scenarios, recently, many nonlinear variants of the IV regression have developed by \citet{1hartford2017deep, 15xu2020dfiv, 22singh2019kernel, 23muandet2019dual} toward complex and entangled causal relationships. To address the challenges from complex nonlinear effects, they firstly learn a joint mapping from the instruments $Z$ and the observed confounders $X$ to the conditional distribution of the treatment $T$ (Fig. 1(b)), i.e., $Pr(T|Z, X)$. This technique introduces higher-order interaction terms between IVs and covariates so that the information from IVs is preserved as much as possible, alleviating the weak instrumental variables problems. 
Then, they estimate treatment $\hat{T}$ from $Pr(T|Z, X)$ obtained in stage 1 and perform nonlinear regression from the estimated treatment $\hat{T}$ and the observed confounders $X$ to the outcome $Y$ in stage 2 (Fig. 1(c)).

From the processes of these nonlinear IV methods (Fig. 1(b \& c)), we know that the confounders $X$ would affect the estimated treatment $\hat{T}$ obtained from stage 1, and also influences the outcome $Y$. Thereby, the distribution of $X$ would become imbalanced
{across different values of $\hat{T}$, that could bias the treatment effect estimation when the outcome model is misspecified.}
{Therefore, explicitly removing confounding bias can be useful for treatment effect estimation.}
Moreover, these methods are reliable if and only if the pre-defined IV is a valid instrument that only affects the outcome through its strong association with 
{the treatment variable}, called exclusion assumption. Nevertheless, since the assumption is untestable in real applications, 
{it can be challenging} to find valid IVs, which hinders the application of the IV-based methods.

In this paper, we focus on the treatment effect estimation with IV regression, and we propose a Confounder Balanced IV Regression (CB-IV\footnote{The code is available at: {https://github.com/anpwu/CB-IV}}) 
algorithm to further remove the confounding bias from observed confounders by balancing in nonlinear scenarios. Specifically, our CB-IV algorithm contains the following three main components: (i) treatment regression: given $Z$ and $X$, identify conditional probability distribution of the treatment variable $T$ (i.e., $Pr(T|Z, X)$) to remove the confounding from unmeasured confounders; 
(ii) confounder balancing: learn a balanced representation of observed confounders $C=f_\theta(X)$, {where $C$ is independent of} the estimated treatment $\hat{T} \sim Pr(T|Z, X)$, to reduce the confounding from $X$ as shown in the Figure~\ref{figure1}(c); and (iii) outcome regression: regressing the outcome $Y$ on the estimated treatment $\hat{T}$ and representation $C$ of observed confounders $X$.

The effectiveness of IV regression algorithms and variants relies on the well pre-defined IVs, however, these algorithms have a major disadvantage that little prior or theory is currently available to find a valid IV in many real applications. To address this problem, compared with the preliminary version \citep{wu2022instrumental}, we extend our CB-IV algorithm to two more general settings without pre-defined IVs, i.e.,  indistinguishable IVs implicitly present in observations, and latent IVs do not appear in observations. Specifically, we extend our CB-IV by a latent-variable module, namely CB-IV-L algorithm, which automatically learns latent proxy from the observations for IVs and confounders. In more general scenarios without pre-defined IVs, our CB-IV-L achieves an accurate treatment effect estimation. 

Theoretically, through redefining and solving the inverse problems under weaker identification assumptions, our CB-IV(-L) algorithm can unbiasedly estimate the treatment effect and achieve lower variance. 
Empirically, we implement extensive experiments and demonstrate the effectiveness of our proposed CB-IV(-L) on different tasks. The main contributions in this paper are as follows:
\begin{itemize}
  \item We study the problem of treatment effect estimation with IV regression, and we find that previous IV regression methods would suffer from the confounding bias from the observed confounders, if the outcome model is misspecified and covariates are imbalanced.
  \item We propose a Confounder Balanced IV regression (CB-IV) method to jointly remove the bias from both the unmeasured confounders with IV regression and the observed confounders by balancing. 
  To the best of our knowledge, this is the first work to combine confounder balancing in IV regression.
  \item Without pre-defined IVs, to two more general settings, i.e.,  indistinguishable IVs implicitly present in observations,and latent IVs do not appear in observations, we extend our CB-IV by a latent-variable module, namely CB-IV-L algorithm, which automatically learns latent proxy from the observations for IVs and confounders. 
  \item In two general settings satisfying homogeneous instrument-treatment association or homogeneous treatment-outcome association, respectively, we give and solve the inverse relationship of counterfactual outcome function from our algorithms. Empirical experiments demonstrate the effectiveness of our algorithms. 
\end{itemize}

\section{Related Works}
\subsection{Instrumental Variable Methods}
A popular way to estimate the causal effect from observational data in the presence of unmeasured confounders is to use an instrumental variable (IV). As a classical IV method, two-stage least squares \citep{16angrist1995identification,17angrist2001instrumental,22222kuang2020ivy} performs linear regression to model the relationship between the treatment and outcome conditional on the instruments. To relax the linearity assumption, nonlinear IV regression variants learn a joint mapping from the instruments $Z$ and observed confounders $X$ to the treatment $T$ in stage 1.
Sieve IV derives a finite dictionary of basis functions to replace the linear counterparts on the structural function and derives a lower bound. \citep{28chen2018optimal,29newey2003instrumental}. Kernel IV \citep{22singh2019kernel} and Dual IV \citep{23muandet2019dual} implement 2-stage regression via mapping $X$ to a reproducing kernel Hilbert space and performing kernel ridge regression. DFIV \citep{15xu2020dfiv} adopts deep neural nets to replace the kernel counterparts. DeepIV \citep{1hartford2017deep} and OneSIV \citep{27lin2019one} estimate the conditional probability distribution of treatment $T$ using the instruments $Z$ and confounders $X$ in stage 1 and perform a joint mapping from resampled treatment $\hat{T} \sim P(T|Z,X)$ and confounders $X$ to the outcome $Y$ in stage 2. 

As shown in Figure~\ref{figure1}(c), if the outcome model is misspecified, imbalanced variables $X$ will bring confounding bias for outcome regression in stage 2 in previous IV regression methods.
To balance the overall sample, \citet{abadie2003semiparametric, singh2019automatic} uses regularized machine learning and achieves semiparametric efficiency with automatic kappa weights, which requires binary instrument, binary treatment and high dimensional covariates.
In general settings, we propose a novel algorithm to combine confounder balance techniques with IV regression. To the best of our knowledge, this is the first provably efficient algorithm that combines the IV method with the confounder balance technique using deep representation learning.

\subsection{Confounder Balance with Representation Learning}
\label{confounding_balance}
Non-experimental studies are increasingly used to estimate treatment effects, and systematic differences between different treatment groups would introduce confounding bias. 
Inspired by traditional confounder balance works \citep{11111kuang2020causal}, such as propensity score methods\citep{31rosenbaum1987model,32li2016matching,33li2020survey}, re-weighting methods\citep{42athey2018approximate,43he2009learning}, Doubly Robust \citep{40funk2011doubly} and backdoor criterion \citep{52pearl2009causal}, CFR \citep{44johansson2016learning,45shalit2017estimating} formulates the problem of confounder balance as a covariate shift problem, and regard the treated group as the source domain and the control group as the target domain for domain adaptive balance under the unconfoundedness assumption. \citet{44johansson2016learning,45shalit2017estimating} expect that representation $C=f_\theta(X)$, from all confounders $X$, discard information related to $T$, but retain as much information related to $Y$ as possible. SITE \citep{51yao2018representation} preserves local similarity and simultaneously balances the representation $C$ distributions. DR-CFR \citep{48hassanpour2019counterfactual, 49hassanpour2019learning} and DeR-CFR \citep{50wu2022learning} propose a disentanglement framework to identify the representation of confounders from all observed variables. 

Deep representation learning has good performance and can capture complex relationships among treatment, observed confounders, and outcome, but it requires the unconfoundedness assumption. Based on these confounder balance methods, we propose to use an instrumental variable to eliminate the bias from the unmeasured confounders.

\subsection{Latent-Variable Modules}
In the presence of unmeasured confounders and no pre-defined IVs, machine learning practitioners turn to the latent-variable model to reconstruct the latent high-level variables and use them to estimate counterfactual outcome \citep{louizos2017cevae,khemakhem2020ivae,zhang2021tedvae,shen2022disentangled}. 
In particular, \citet{louizos2017cevae} regards the observations as noisy proxy variables for unobserved confounders and proposes a causal effect variational autoencoder (CEVAE) for recovering causal effect and performing causal inference. 
Inspired by this, TEDVAE \citep{zhang2021tedvae} differentiates confounding factors
from instrumental and risk factors for treatment effect estimation and disentangles the
latent factors into three corresponding sets. $\beta$-Intact-VAE \citep{wu2022betaintactvae} recovers a prognostic score using a variational autoencoder, thereby mapping a higher dimensional set of covariates with limited overlap to a lower dimensional set where overlap holds, and such that ignorability is maintained.

The above methods are derived from CEVAE, and also inherit its disadvantages.
\citet{rissanen2021criticalcevae} investigate this gap between theory
and empirical results on CEVAE: it seems to work reliably under simple scenarios (e.g., only one unmeasured confounder), and does not accurately recover all confounders with a complex data distribution or high-dimensional unmeasured variables, as opposed to its original motivation. 
As an alternative, we propose a latent-variable module to recover the part of unmeasured IVs instead of full confounders or full IVs and plug it into the treatment regression model.

\section{Problem Setting and Assumptions}

In this paper, we aim to estimate the average treatment effect by the structural function from observational data in the presence of unmeasured confounders. In the observational data $\mathbb{D}=\{z_i,x_i,t_i,y_i\}_{i=1}^n$, for each unit $i$, we observe a treatment variable $t_i \in T$ where $T \subset  \mathbb{R}$, a outcome variable $y_i \in Y$ where $Y \subset \mathbb{R}$, 
instrumental variables $z_i \in Z$ where $Z \subset  \mathbb{R}^{m_Z}$,  and observed confounders $x_i \in X$ where $X \subset  \mathbb{R}^{m_X}$. Besides, there are some unmeasured confounders $u_i \in U$ where $U \subset  \mathbb{R}^{m_U}$ and might affect or be affected by $x_i$, but not recorded in the observational data.
$m_X, m_Z$ and $m_U$ are the dimensions of the observed confounders $X$, instrumental variables $Z$ and unobserved confounders $U$.
The causal relationship can be represented with the following model (as shown in Figure~\ref{figure1}(a)):
\begin{align}
   {\{Z,X,U\} \rightarrow T; \{T,X,U\} \rightarrow Y; Z \perp U,X; X \not \perp U }
\end{align}
Nevertheless, due to the untestable exclusion assumption, no pre-defined IVs would be more common in real-world scenarios, e.g., indistinguishable IVs implicitly present in observations, and latent IVs do not appear in observations. The former is referred to as the mixed-variable setting, we extend our CB-IV by using all the observed covariates as input to implement two-stage regression, in Section~\ref{sub:CB-IV}
We refer to the latter as a latent-variable setting, i.e., no valid IVs presented in the observations, and we will elaborate a variational inference model to learn latent IVs and confounders, and extend our CB-IV by a latent-variable module, namely CB-IV-L algorithm, in Section~\ref{sub:CB-IVl}.

\begin{definition}
\textbf{The average treatment effect} ($ATE$): 
\begin{align}
    {ATE(t) = \mathbb{E}[Y \mid {do}(T=t), X] - \mathbb{E}[Y \mid {do}(T=0), X]}
\end{align}
\end{definition}
\begin{definition}
\textbf{An Instrument Variable $Z$} is an exogenous variable that affects the treatment $T$, but does not directly affect the outcome $Y$. Besides, an valid instrument variable satisfies the following three assumptions: \\
\textbf{Relevance:} $Z$ is a cause of $T$, i.e., $Pr(T \mid Z) \neq Pr(T)$. \\
\textbf{Exclusion:} $Z$ does not directly affect the outcome $Y$, i.e., $Z \perp Y \mid T, X, U$.  \\
\textbf{Unconfounded:} $Z$ is independent of all confounders, including $X$ and $U$, i.e., $Z \perp X,U$
\end{definition}

\noindent \textbf{Identification:} Even if the instrument satisfies these assumptions, at least one of the two homogeneity assumptions is required to identify the average treatment effect of $T$ on $Y$ \citep{16angrist1995identification,angrist1996identification,29newey2003instrumental, 53hernan2010causal,56wooldridge2010econometric}. The identifying assumptions in our paper basically follow the homogeneity assumptions \citep{heckman2006understanding,hernan2006instruments,hartwig2020average}, which is a more general version than Monotonicity Assumption \citep{16angrist1995identification,angrist1996identification} and Additive Noise Assumption \citep{29newey2003instrumental}) in the econometrics literature \citep{56wooldridge2010econometric,hartwig2020average,hartwig2021homogeneity}. The two homogeneity assumptions are as follows: 

\begin{assumption}
\label{assumZT}
\textbf{Homogeneous Instrument-Treatment Association}: The association between the IV and the treatment is homogeneous in the different levels of unmeasured confounders, i.e., $\mathbb{E}[T|Z=a,  U]-\mathbb{E}[T|Z=b, U]=\mathbb{E}[T|Z=a]-\mathbb{E}[T|Z=b]$.
\end{assumption}

\begin{assumption}
\label{assumTY}
\textbf{Homogeneous Treatment-Outcome Association}: The association between the treatment and the outcome is homogeneous in the different levels of unmeasured confounders, i.e., $\mathbb{E}[Y|T=a,  U]-\mathbb{E}[Y|T=b, U]=\mathbb{E}[Y|T=a]-\mathbb{E}[Y|T=b]$.
\end{assumption}

\noindent \textbf{Discussion about Confounder Imbalance:} To precisely estimate the treatment effect, The implementation of IV methods estimates a conditional treatment distribution ${Pr}(T \mid Z, X)$ using $\{Z,X\}$ in the treatment regression stage, then learns \textbf{the counterfactual outcome function} $h(T,X)$ from the re-sampled treatment $\hat{T} \sim P(T|Z,X)$ and the variables $X$ to $Y$ directly: 
\begin{eqnarray}
\label{counterfactual}
    h(\hat{T},X) = \mathbb{E}[Y \mid \hat{T}, X].
\end{eqnarray}
In the outcome regression stage, the confounders $X$ would affect the resampled treatment $\hat{T}$ obtained from stage 1 as shown in the Figure~\ref{figure1}(b), leading to an imbalance of $X$ between different resampled treatment options in stage 2 since the lack of randomization \citep{Confounder1:cook2002experimental}. 
If the outcome model is misspecified, 
such confounder imbalance would bring confounding bias for outcome regression in the previous IV-based methods, especially with high dimensional $X$, introducing bias and large variance in the estimation \citep{Confounder3:schroeder2016propensity}, i.e., $h(T,X) = \mathbb{E}[Y \mid T, X]$ holds only in the same distribution, and $h(t,X) \not = \mathbb{E}[Y \mid {do}(T=t), X]$ out of the distribution. 

Based on this, we firstly propose CB-IV to reduce the confounding bias from observed confounders by confounder balancing (details in Section~\ref{sec:algorithm}  \& \ref{sec:optimization}) in the outcome regression and re-build the inverse problem (details in Section~\ref{sec:inverse}) for relationship for the counterfactual prediction function $h(T,X)$. Moreover, without pre-defined IVs, we propose a variational inference model to learn latent IVs and confounders, and extend our CB-IV to CB-IV-L algorithm (details in Section~\ref{sec:extend}).

\section{Confounder Balancing for Instrumental Variable Regression}

In this section, we first introduce the proposed algorithm (CB-IV) and achieve balanced confounder representation for eliminating confounding bias in Section~\ref{sec:algorithm}; then, we provide a two-stage optimization for CB-IV, and elaborate the network structures and pseudo-code of CB-IV in Section~\ref{sec:optimization}; with representation obtained from CB-IV algorithm, we re-identify the inverse relationship for counterfactual outcome function and avoid ill-posed identification problem under general settings in Section~\ref{sec:inverse}. 
The results in Section~\ref{sec:inverse} can justify that the proposed CB-IV algorithm can achieve an accurate and robust estimation.

\subsection{Algorithm}
\label{sec:algorithm}

IV regression is the classical method for addressing the unmeasured confounders, but recent nonlinear IV-based methods suffer the confounding bias from the observed confounders, as shown in the Figure~\ref{figure1}(c), leading to poor performance on outcome regression and treatment effect estimation in practice.
To address these challenges, we propose a Confounder Balanced IV Regression (CB-IV) algorithm to achieve confounder balancing in IV regression. Specifically, confounder balancing for removing the bias from observed confounders and IV regression for eliminating the bias from unmeasured confounders. The proposed CB-IV algorithm consists of three main components: (i) treatment regression, (ii) confounder balancing, and (iii) outcome regression. Next, we would
elaborate CB-IV algorithm for binary treatment cases and multi-value \& continuous treatment cases, respectively. 

\subsubsection{Treatment Regression}
\textbf{For binary treatment cases}, we regress treatment $T$ with IVs $Z$ and observed confounders $X$ directly, as the treatment regression stage did in the previous nonlinear IV-based method. Specifically, we estimate the conditional probability distribution of the treatment $\hat{Pr}(T|Z, X)$ with a logistic regression network $\pi_\omega(z_i, x_i)$ with learnable parameter $\omega$ for each unit $i$, and optimize the following loss function for treatment regression:
\begin{eqnarray}
\label{eq:loss_T}
\mathcal{L}_T = -\frac{1}{n} \sum_{i=1}^n ( t_i \log{ \left(\pi_\omega(z_i, x_i)\right)} + (1-t_i) \left(1-\log{ \left(\pi_\omega(z_i, x_i)\right)}\right)).
\end{eqnarray} 
In subsequent sections, we use $\hat{Pr}(T=1|Z, X) = \pi_\omega(z_i, x_i)$ and $\hat{Pr}(T=0|Z, X) = 1-\pi_\omega(z_i, x_i)$ to denote the probability distribution estimation. 

\noindent \textbf{For multi-value \& continuous treatment cases}, we model the conditional probability distribution, $\hat{P}(T|Z, X) \sim \mathcal{N}(\mu_{\omega}(Z,X),\sigma_{\omega}(Z,X))$ with neural networks $\{\mu_{\omega}, \sigma_{\omega}\}$, using a mixture of gaussian distribution with multiple sub-networks $\{\mu_{\omega,k}, \sigma_{\omega,k}\},k=1,...,K$, where $K$ denotes the number of latent gaussian distributions. We sample $m$ (the larger the better) treatment $\{\hat{t}_i^j\}_{j=1,...,m} $ for each unit $\{z_i,x_i\}$ to approximate the true treatment $t_i$: $\Scale[1.0]{\mathcal{L}_T = \frac{1}{nm} \sum_{i=1}^{n}\sum_{j=1}^{m} \left(t_i - \hat{t}_i^j\right)^2}$, where $\hat{t}_i^j \sim \mathcal{N}(\mu_{\omega}(Z,X),\sigma_{\omega}(Z,X))$. As the number of samples $m$ increases, the above objective will converge to the point estimation:
\begin{eqnarray}
\label{eq:loss_Tcont2}
\mathcal{L}_T = \frac{1}{n} \sum_{i=1}^{n} \left(t_i - \hat{t}_i\right)^2, \hat{t}_i = \mu_{\omega}(z_i,x_i).
\end{eqnarray} 
In subsequent sections, we use $\hat{T} = \mu_{\omega}(Z,X)$ to denote the estimated treatment conditional on the observed instruments and covariates.

\subsubsection{Confounder Balancing}
After treatment regression, we can obtain the causal graph as shown in Figure~\ref{figure1}(c), where the observed variables $X$ would become the confounders for outcome regression. To address this problem, we propose to learn a balanced representation of observed confounders $C=f_\theta(X)$, independent with the estimated treatment $\hat{T} \sim Pr(T|Z, X)$, to reduce the confounding from $X$.

\noindent \textbf{For binary treatment cases}, 
we learn the balanced representation $C=f_\theta(X)$ by minimizing the discrepancy of distributions for different treatment arms to achieve $C \perp \hat{T}$ for confounder balancing: 
\begin{eqnarray}
\mathcal{L}_C=\text{ disc}(\hat{T}, f_\theta(X)) =  \text{IPM} (\{f_\theta(x_i)\hat{Pr}(t_i=0 \mid z_i, x_i)\}_{i=1}^n, \{f_\theta(x_i)\hat{Pr}(t_i=1 \mid z_i, x_i)\}_{i=1}^n),
\label{wass-cont}
\end{eqnarray}
where $\{f_\theta(x_i)\hat{Pr}(t_i=k \mid z_i, x_i)\}_{i=1}^n, k \in \{0,1\}$ denotes the distribution of representation $C=f_\theta(x_i)$ in the group $T=k$ given the $\hat{Pr}(t_i \mid z_i, x_i)$. In this paper, we choose Wass distance as the integral probability metric (IPM) to measure the discrepancy of distributions.

As a distance measure widely used in deep learning, Wasserstein distance (Wass) has consistent estimators which can be efficiently computed in the finite sample case \citep{45shalit2017estimating,67sriperumbudur2012empirical}. In causal inference, practitioners tend to adopt Wasserstein distance (Wass) to calculate the dissimilarity of distributions from different treatment arms and fit a balanced representation by minimizing the discrepancy \citep{45shalit2017estimating,49hassanpour2019learning}. Hence, we also choose it as the measurement metric in binary treatment cases. 

\noindent \textbf{For multi-value \& continuous treatment cases},
we learn a balanced representation (i.e., $C$) of the observed confounders $X$ as $C=f_\theta(X)$ via mutual information (MI) minimization constraints \citep{60cheng2020club}: firstly, we use variational distribution $Q_{\psi}(\hat{T} \mid C) = \mathcal{N}(\mu_{\psi}(C),\sigma_{\psi}(C))$ parameterized by neural networks $\{\mu_{\psi}, \sigma_{\psi}\}$ to approximate the true conditional distribution $P (\hat{T} \mid C)$; then, we minimize the log-likelihood loss function of variational approximation $Q_{\psi}(\hat{T} \mid C)$ with $n$ samples to estimate MI:
\begin{eqnarray}
{\mathcal{L}_C=\text{ disc}(\hat{T}, f_\theta(X)) =\frac{1}{n^2}\sum_{i=1}^{n}\sum_{j=1}^{n}\left[\log Q_{\psi}\left(\hat{t}_{i}\mid c_{i}\right)-\log Q_{\psi}\left(\hat{t}_{j}\mid c_{i}\right)\right]}, C=f_\theta(X). 
\end{eqnarray}
We adopt an alternating training strategy to iteratively optimize $Q_{\psi}(\hat{T} \mid C)$ and the network $C=f_{\theta(X)}$ to implement balanced representation in the Confounder Balancing. 

\subsubsection{Outcome Regression}
Finally, we propose to regress the outcome with the estimated treatment $\hat{T} \sim P(T|Z, X)$ obtained in the treatment regression module and the representation of confounders $C=f_\theta(X)$ obtained in confounder balancing module. 

\noindent \textbf{For binary treatment cases}, 
considering that high dimensional representation $f_\theta(X)$ would induce the loss of binary treatment information in outcome regression function $h_\xi(\hat{T},f_\theta(X))$ \citep{45shalit2017estimating} with single neural network $h_\xi(\cdot)$. We propose to  regress the potential outcome (i.e., $Y(do(T=1),X)$ and $Y(do(T=0),X)$) by optimizing $h_{\xi^0}(f_\theta(X))$ and $h_{\xi^1}(f_\theta(X))$ as two different regression network with learnable parameters $\xi^0$ and $\xi^1$, respectively:
\begin{eqnarray}
\mathcal{L}_Y=\Scale[1.0]{\frac{1}{n} \sum\limits_{i=1}^n \left
(y_i - \sum\limits_{t_i \in \{0,1\}} h_{\xi^{t_i}}(f_\theta(x_i)) \hat{Pr}(t_i \mid z_i, x_i)\right)^2},
\label{tenfour}
\end{eqnarray}
where $\hat{Pr}(t_i = 1 \mid z_i, x_i) = \pi_\omega(z_i, x_i)$ and $f_\theta(x_i)$ are derived from the treatment regression module and the confounder balancing module, respectively.

\noindent \textbf{For multi-value \& continuous treatment cases}, 
we regress the outcome with the estimated treatment $\hat{T} = \mu_{\omega}(Z,X)$ and the representation of confounders $C=f_\theta(X)$ directly:  
\begin{eqnarray}
\mathcal{L}_Y=\Scale[1.0]{\frac{1}{n} \sum\limits_{i=1}^n \left
(y_i - h_\xi(\hat{t}_i,f_\theta(x_i))\right)^2},
\label{tenfourcont}
\end{eqnarray}
where $\hat{t}_i \sim \hat{P}(T|Z, X)$ and $f_\theta(x_i)$ are derived from the treatment regression module and confounder balancing module, respectively. In subsequent sections, we use $h_{\xi^{t_i}}(f_\theta(x_i))$ to represent $h_\xi(t_i,f_\theta(x_i))$ for simplicity.

\subsection{Optimization and Network Structures} 
\label{sec:optimization}

\subsubsection{Optimization} 
Like the optimization of the previous IV regression method, we perform a two-stage optimization for our CB-IV, which applies to both binary, multi-value and continuous cases. 
In stage 1, we optimize the treatment regression module $\pi_\omega$ by minimizing the loss $\mathcal{L}_T$ defined in Eq. (\ref{eq:loss_T}) or Eq. (\ref{eq:loss_Tcont2}):
\begin{eqnarray}
\label{loss1}
\min_{\omega} \mathcal{L}_T.
\end{eqnarray}
In stage 2, then, we simultaneously optimize the confounder balancing and outcome regression modules by setting the balanced learning representations $C$ as a kind of regularization on the outcome regression model, with the following loss:
\begin{eqnarray}
\label{loss2}
\min_{\theta, \xi} \mathcal{L}_Y +\alpha \mathcal{L}_C,
\end{eqnarray}
where $\alpha$ is a trade-off hyper-parameter. In mathematical, our optimal objective (Eq. (\ref{loss2})) is consistent with the error bound developed by \citep{45shalit2017estimating}. The expected treatment effect estimation error $\epsilon(h, \theta)$ can be bounded by the standard generalization error and the distance between the treated and control distributions induced by the representation: 
\begin{eqnarray}
    \epsilon(h, \theta) \leq 2\left( \int_{\mathcal{T}} \epsilon_{F}^{T=t}(h, \theta) d t+B_{\theta} \text{ disc}(\hat{T}, f_\theta(X))-2 \sigma_{Y}^{2}\right),
\end{eqnarray}
where $\epsilon_{F}^{T=t}(h, \theta)=\int_{\mathcal{X}} \ell_{2}(y, h(T=t,f_\theta(x))) p^{T=t}(x) d x$ and $p^{T=t}(x)$ denotes the PDF of $x$ given $T=t$, in the observational data; $B_{\theta}$ is a constant; $ \sigma_{Y}^{2}$ is the expected variance of the outcome $Y$. If we directly regress $\mathbb{E}[Y|T,X]=h_\xi(T,X)$, non-parametric models without prior knowledge may have poor performance for rare samples in training data (overfitting). Thus, confounder balance is a great regularization of the outcome regression model.

Then, the average treatment effect can be estimated by
\begin{eqnarray}
\label{ATEY}
\widehat{ATE}(t) = \mathbb{E}[h_{\xi^t}(f_\theta(X)) - h_{\xi^0}(f_\theta(X))]. 
\end{eqnarray}

\subsubsection{Network Structures} 

\begin{table}[t]
\centering
\caption{Network Structures of CB-IV on Synthetic and Semi-Synthetic Datasets.}
    \scalebox{1.0}{
    \begin{tabular}{c | c c c c c}
        \hline
         {\textbf{Stage}} & {\textbf{Setting}} &  {\textbf{Syn}} &  {\textbf{Demand}} &  {\textbf{IHDP}} &  {\textbf{Twins}} \\
        \hline
        \multirow{5}[8]{*}{\textbf{\shortstack{Treatment \\ Regression}}} 
          & Loss & log & MSE & log & log  \\
          & Epoch & 3 & 20 & 3 & 3  \\
          & Batchsize & 500 & 500 & 500 & 500 \\
          & MLPLayers & [128,64] & [128,64] & [128,64] & [128,64] \\
          & Activation & ReLU & ReLU & ReLU & ReLU \\
          & BatchNorm & True & True & True & True \\
          & Learning\_Rate & 0.05 & 0.005 & 0.05 & 0.05 \\
          & Optimizer & SGD & SGD & SGD & SGD \\
        \hline
        \multirow{6}[9]{*}{\textbf{\shortstack{Outcome \\ Regression}}} 
          & Loss & MSE & MSE & MSE & log  \\
          & Epoch & 3000 & 6000 & 100 & 200  \\
          & Batchsize & 256 & 200 & 100 & 100 \\
          & MLPLayers\_R & [256]$*$3 & [256]$*$3 & [200]$*$3 & [256]$*$3 \\
          & MLPLayers\_Y & [256]$*$5 & [256]$*$5 & [100]$*$3 & [128]$*$5 \\
          & Activation & ELU & ELU & ELU & ELU \\
          & BatchNorm & False & False & False & False \\
          & Learning\_Rate & 0.0005 & 0.005 & 0.0005 & 0.0005 \\
          & Optimizer & Adam & Adam & Adam & Adam \\
          & $\alpha$ & 0.01/0.001 & 0.1 & 0.1 & 0.001/0.0001 \\
         \hline
    \end{tabular}
    }
    \label{param-detail}%
\end{table}%

For the Treatment Regression, we use multi-layer perceptrons with ReLU activation function and BatchNorm as our logistic regression network $\pi_\omega$, and the network has two hidden layers with 128, 64 units, respectively. Then, We use stochastic gradient descent (SGD, \citep{62duchi2011adaptive}) to train the network $\pi_\omega$ with a loss $\mathcal{L}_T$ for three epochs with batch size 500. 

For the Outcome Regression and Confounder Balancing, we use Adam \citep{63kingma2014adam} to train the networks $f_\theta, h_{\xi^{t}}$ with the loss $\mathcal{L}_Y+\alpha \mathcal{L}_C$ jointly. To prevent overfitting, we regularize the prediction functions $h_{\xi^{t}}$ with a small $l_2$ weight decay as a regularization.

Table~\ref{param-detail} lists the optimal structure networks of CB-IV used
for each dataset in the paper’s experiments.
In the Treatment Regression Stage, the loss would be an MSE-loss for continuous treatment and a log-loss for binary treatment, and the treatment network has multiple hidden layers with [MLPLayers] units. In the Treatment Regression Stage, the loss would be an MSE-loss for continuous outcome and a log-loss for binary outcome. The representation network has multiple hidden layers with [MLPLayers\_R] units, and the outcome network has multiple hidden layers with [MLPLayers\_Y] units. Algorithm \ref{algorithm} shows the pseudo-code of our CB-IV for binary treatment cases. It can be easily changed to multi-value treatment and continuous treatment.  

Hardware used: Ubuntu 16.04.5 LTS operating system with 2 * Intel Xeon E5-2678 v3 CPU, 384GB of RAM, and 4 * GeForce GTX 1080Ti GPU with 44GB of VRAM.
    
Software used: Python with TensorFlow 1.15.0, NumPy 1.17.4, and MatplotLib 3.1.1.

\begin{algorithm}[t]
    \caption{CB-IV (binary treatment as a show case)}
    \label{algorithm}
    \begin{algorithmic}
	\STATE \textbf{Input:} Observational data $\mathbb{D}=\{z_i,x_i,t_i,y_i\}_{i=1}^n$; Maximum number of iterations $\mathcal{I}$
	\STATE \textbf{Output:} $\hat{Y}_0=h_{\xi^0}(f_\theta(X)), \hat{Y}_1=h_{\xi^1}(f_\theta(X))$
	\STATE \textbf{Loss function:} $\mathcal{L}_T$ and $\mathcal{L}_Y+\alpha \mathcal{L}_C$
	\STATE \textbf{Components:} Logistic regression network $\pi_\omega(\cdot)$; Representation learning network $f_\theta(\cdot)$; Two-head outcome regression networks $h_{\xi^0}(\cdot)$ and $h_{\xi^1}(\cdot)$. 
	\STATE \textbf{Treatment Regression Stage:}
	\FOR{$\text{itr}=1$ {\bfseries to} $\mathcal{I}$}
	\STATE $\{z_{i}, x_{i}\}_{i=1}^n \rightarrow \pi_\omega(z_{i}, x_{i}) \rightarrow \hat{P}(t=1 \mid z_i, x_i)$
	\STATE $\mathcal{L}_T = -\frac{1}{n} \sum_{i=1}^n \left( t_i \log{ \left(\pi_\omega(z_i, x_i)\right)} + (1-t_i) \left(1-\log{ \left(\pi_\omega(z_i, x_i)\right)}\right)\right)$
	\STATE update $\omega \leftarrow {\rm{SGD}}\{\mathcal{L}_T\}$
	\ENDFOR
	\STATE \textbf{Outcome Regression Stage:}
	\FOR{$\text{itr}=1$ {\bfseries to} $\mathcal{I}$}
	\STATE $\left\{x_{i}\right\}_{i=1}^{n} \rightarrow C_i=f_\theta(x_i)$
	\STATE $\{z_{i}, x_{i}\}_{i=1}^n \rightarrow \pi_\omega(z_{i}, x_{i}) \rightarrow \hat{P}(t=1 \mid z_i, x_i)$
	\STATE $\left\{f_\theta(x_i), t_{i}\right\}_{i=1}^{n} \rightarrow \text{disc}(\hat{T},f_\theta(X))$
	\STATE $\mathcal{L}_Y+\alpha \mathcal{L}_C =  \frac{1}{n} \sum_{i=1}^n \left(y_i - \sum_{\hat{t} \in \{0,1\}} h_{\xi^{\hat{t}}}(f_\theta(x_i)) \hat{P}(\hat{t} \mid z_i, x_i)\right)^2 + \alpha \text{ disc}(\hat{T}, f_\theta(X))$
	\STATE update $\theta, \xi^0, \xi^1 \leftarrow {\rm{Adam}}\{\mathcal{L}_Y+\alpha \mathcal{L}_C\}$
	\ENDFOR
\end{algorithmic}
\end{algorithm}

\subsection{Inverse Problem for Counterfactual Outcome Function}
\label{sec:inverse}

Recent IV methods \citep{1hartford2017deep,29newey2003instrumental,27lin2019one} define an inverse problem for the counterfactual prediction function $h(T,X)$ with two observable functions $\mathbb{E}[Y \mid T, X]$ and $P(T\mid Z,X)$:
\begin{eqnarray}
\label{inverseTX}
    \mathbb{E}[Y \mid Z, X] = \int h(T,X) d {P}(T \mid Z, X).
\end{eqnarray}
The inverse relationship for ${ h  }(T,X)$ holds only under \textit{the additive noise assumption} on counterfactual outcome function:
\begin{eqnarray}
\label{additivenoise}
Y = g(T,X)+U, \mathbb{E}[U \mid Z]=0 .
\end{eqnarray}
Nevertheless, the outcome functions are agnostic and cannot be artificially controlled and assumed. In contrast, the treatment made by human decision-making is always traceable. For example, consider a promotional activity that will affect the buying tendency of people homogeneously, but it is not easy to discuss the impact on the employment rate of these people in the future. Thus, we believe Homogeneous Instrument-Treatment Association \citep{hartwig2020average, hartwig2021homogeneity} is a more common setting in the real world. Based on the Homogeneous Instrument-Treatment Assumption, we model a more general causal relationship by relaxing the additive assumption to the multiplicative assumption on counterfactual outcome function as:
\begin{align}
\label{complicated_function1}
    & T = f_1(Z,X)+f_2(X,U) \\ 
    & Y = g_1(T,X)+g_2(T)g_3(U)+g_4(X,U), Z \perp U,X
\label{complicated_function2}
\end{align}
where ${f_{i}}(\cdot),{g_{j}}(\cdot)$ are unknown and potentially nonlinear continuous functions. $g_2(T)g_3(U)$ denotes the multiplicative terms of $U$ with $T$ (e.g., $U^2T-UT+U$), and we define it as \textit{the multiplicative assumption}. 
Due to the ill-posed inverse problem, traditional IV approaches rarely discuss homogeneity assumptions. This paper redefines an inverse problem by introducing balanced representations $C$. The completeness of $Pr({T \mid Z,X})$ and $Pr({Y \mid T,X})$ guarantees the uniqueness of the solution \citep{29newey2003instrumental}. 

The Eqs. (\ref{complicated_function1}) \& (\ref{complicated_function2}) are a general form of the homogeneity assumptions, and \textbf{the Counterfactual Prediction Function} can be re-defined as:
\begin{eqnarray}
    \nonumber &&  \mathbb{E}[Y \mid \text{do}(T), C, X] =  \mathbb{E}[{ h  }(T,C) \mid T,C,X] \\
    & = &   \mathbb{E}\left[g_1^C(T,C) + g_2(T)\mathbb{E}[g_3(U) \mid C \right] +  \mathbb{E}\left[g_4(X,U) \mid C] \mid T,C,X \right] ,
\end{eqnarray}
where ${ h  }(T,C)$ is the conditional expectation of $Y$ given the observables $T$ and disentangled representation $C$. We transform $g_1(T,X)$ as $g_1^C(T,C)$ with the disentangled representation $C=f_\theta(X)$, satisfying $\mathbb{E}[g_1^C(T,C) \mid T,C,X ] = \mathbb{E}[g_1(T,X) \mid T,C,X ]$. $\mathbb{E}[g_3(U) \mid C]$ and $\mathbb{E}[g_4(X,U) \mid C]$ are constant for the specified $C$. 

Under the causal relationship (\ref{complicated_function1}) \& (\ref{complicated_function2}), we show the inverse problem for the counterfactual outcome function, implying that the identification of the counterfactual prediction function can be identified, as follows:

\begin{theorem}
\label{theory_1} \textbf{Inverse Relationship of Eqs. (\ref{complicated_function1}) \& (\ref{complicated_function2})}. If the learned representation of observed confounders $C=f_\theta(X)$ is independent with the estimated treatment $\hat{T}$, then the counterfactual prediction function ${ h  }(T,C)$ can be identified with instrumental variables $Z$ and representation $C$. Then, we can establish an inverse relationship for ${ h  }(T,C)$ given $\mathbb{E}[Y \mid Z,C,X]$ and $P(T \mid Z,X)$, as follow:
\begin{eqnarray}
\mathbb{E}[Y \mid Z,C,X] = \int \left[ { h  }(T,C) \right] dP(T \mid Z,X),
\end{eqnarray}
where, $dP(T \mid Z,X)$ is the conditional treatment distribution. The proof is deferred to Appendix~\ref{app:theorem}.
\end{theorem}

Based on our proposed counterfactual prediction function $h(T,C)$ with the balanced representation $C$, similarly, we can also establish the inverse relationship of Eq. (\ref{additivenoise}) as:
\begin{eqnarray}
\label{inverseYU}
\mathbb{E}[Y \mid Z, C, X] = \int h(T,C) d {P}(T \mid Z, X),
\end{eqnarray}
where $\mathbb{E}[h(T,C) \mid Z,C,X] = \mathbb{E}[h(T,X) \mid Z,X]$, which is consistent with Eq. (\ref{inverseTX}) under Assumption~\ref{assumTY}. 

\begin{remark}
Sufficient assumptions for identification of average treatment effect \citep{16angrist1995identification,29newey2003instrumental, 53hernan2010causal,hartwig2020average} with the instruments include: homogeneity in the causal effect of $T$ on $Y$ or homogeneity in the association of $Z$ with $T$. 
Due to the ill-posed inverse problem \citep{kress1989linear,29newey2003instrumental}, traditional IV approaches rarely discuss homogeneity assumptions. In this paper, we redefine an inverse problem by introducing balanced representations $C$. The completeness of $Pr({T \mid Z,X})$ and $Pr({Y \mid T,X})$ guarantees the uniqueness of the solution $h(T,C)$ \citep{29newey2003instrumental}. Combining confounder balancing in IV methods, CB-IV redefines and solves the inverse problem under Assumptions~\ref{assumZT} or~\ref{assumTY}, and achieves a more accurate and robust estimation. 
\end{remark}

\section{Two More Challenging Settings without Pre-defined IVs}
\label{sec:extend}

In the above section, we propose a Confounder Balanced IV regression (CB-IV) method to jointly remove the bias from both the unmeasured confounders with IV regression and the observed confounders by balancing. Nevertheless, due to the untestable exclusion assumption, the thorny problem is that we do not have the prior knowledge to distinguish IVs from observed covariates \citep{49hassanpour2019learning,50wu2022learning}, referred to as Mixed-Variable Challenge. 
Besides, unmeasured IVs are a common problem in IV regression algorithms, i.e., no pre-defined IVs for IV regression. We refer to these unmeasured IVs in the data generating process as latent variables, and use variational inference models to learn latent IVs and confounders. We call this problem as Latent-Variable Challenge.

Compared with the preliminary version \citep{wu2022instrumental}, we extend the CB-IV by changing the input to Mixed-Variable Challenge under a weak assumption in Section~\ref{sub:CB-IV}.
In addition, we will extend our CB-IV by a latent-variable module, namely CB-IV-L algorithm, to learn latent IVs and confounders for Latent-Variable Challenge in Section~\ref{sub:CB-IVl}.

\begin{figure}
\begin{center}
\includegraphics[width=0.95\linewidth]{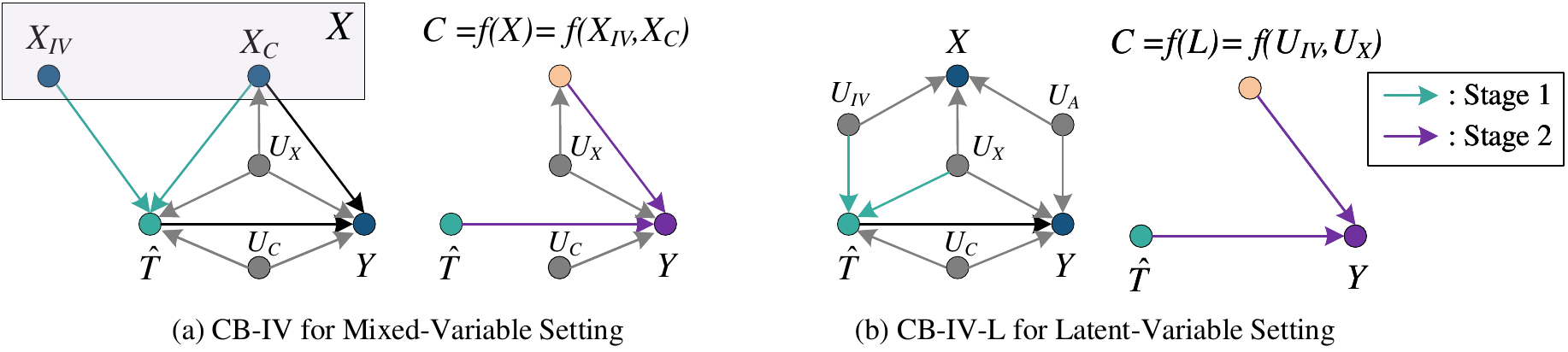}
\end{center}
\vspace{-12pt}
\caption{(a) CB-IV for Mixed-Variable Setting; (b) CB-IV-L for Latent-Variable Setting. Blue nodes denote observable variables, and gray nodes indicate unmeasured variables. Green nodes denote the estimated treatment from stage 1, purple and yellow nodes indicate the outcome of interest and balanced representation in stage 2. Besides, the gray lines encode effects from unmeasured variables. The complete unmeasured confounders are defined as $U=\{U_X, U_C\}$. $U_{IV}$ and $U_{A}$ denotes the unmeasured IVs and adjustments. }
\label{figure2}
\vspace{-15pt}
\end{figure}

\subsection{CB-IV for Mixed-Variable Challenge}
\label{sub:CB-IV}
Due to the untestable exclusion assumption, it is common that we don't have the prior knowledge to distinguish IVs from observed covariates in observational data, i.e., mixed variables. In the presence of unmeasured confounders and no known IVs, we extend CB-IV to mixed variables and use the mixed data directly to perform a two-stage regression with a weak assumption:
\begin{assumption}
\textbf{Mixed-Variable Setting}: We assume, in observational data, we obtain a large number of covariates, and at least one is a valid IV. However, we don't have the prior knowledge to identify which is the IV, i.e., IVs are implicitly present in mixed data. 
\end{assumption}

Under the above assumption, we use the notations $X_{IV}$ and $X_{C}$ to represent IVs and confounders implicitly presented in mixed variables, i.e., $X = \{X_{IV}, X_{C}\}$. Then, we use the mixed variables directly to perform a two-stage regression for CB-IV, as shown in Figure~\ref{figure2}(a): \\
\textbf{In treatment regression stage}, CB-IV concatenates the instruments and confounders into a vector and then uses it to regress the treatment. This means that we can directly replace this concatenation vector with the mixed variables $X = \{X_{IV}, X_{C}\}$, and then perform the same treatment regression operation to obtain estimated treatment $\hat{T}$. \\
\textbf{In outcome regression stage}, CB-IV learns a balanced representation $C=f(X)$, as well as regresses the outcome on the estimated treatment $\hat{T}$ and the representation $C=f(X)$, i.e., $\hat{Y} = h(\hat{T}, C)$ and $\hat{T} \perp C$. In the confounder balancing module, we note that CB-IV will force the representation $C=f(X)$ to balance out the information related to the treatment $T$. By minimizing the mutual information $\text{disc}(\hat{T},f_\theta(X))$, we can ensure the information of the IVs $X_{IV}$ would not be embedded into representation $C=f(X)$, since the instruments $X_{IV}$ is associated with the treatment $\hat{T}$, i.e., $C=f(X_{C})=f(X_{IV}, X_{C})$. This means that we can use the mixed variables $X = \{X_{IV}, X_{C}\}$ directly as input to learning balanced representations $C=f(X)$, since the information from IVs will be automatically filtered. 

Thus, all we need to do is replace the inputs to the treatment regression and confounder balancing modules of CB-IV with the mixed variables $X = \{X_{IV}, X_{C}\}$, and then we can perform the two-stage regression described above to estimate the unbiased treatment effects.

\subsection{CB-IV-L for Latent-Variable Challenge}
\label{sub:CB-IVl}

A more significant challenge is that no valid IVs are included in the observational data. Without any structural assumptions, the treatment effect would be unidentifiable \citep{16angrist1995identification, 29newey2003instrumental, hartwig2020average, hartwig2021homogeneity}. For instance, if all observed and unobserved variables are confounders that are strongly correlated with the treatment variables, then it is not possible to obtain unbiased treatment effect estimation by instrument variable methods. 
In the presence of unmeasured confounders and no pre-defined IVs  observed in data, we extend CB-IV by a latent-variable module, namely CB-IV-L algorithm, to learn latent IVs and confounders with a weak assumption:

\begin{assumption} \label{ass:latent}
\textbf{Latent-Variable Setting}: 
We assume, without loss of generality, that any dataset of the form $\{X, T, Y\}$ is generated from four types of latent factors $\{ U_{IV},  U_X, U_A, U_C \}$, as illustrated in Figure~\ref{figure2}(b). $X$ is the observed variables that may contain noisy proxy variables for latent factors $\{ U_{IV}, U_X, U_A\}$ but not for $U_C$, or itself. For the causal effect of $T$ on $Y$, $U_{IV}$ are latent IVs that affect only the treatment, $U_A$ are latent adjustments that affect only the outcome, and $U_X$ are latent confounders associated with $X$. Note that $U_C$ are latent confounders independent of $X$, i.e., it can not be recovered from $X$. The complete unmeasured confounder $U$ are composed of $\{U_X,U_C\}$, i.e., $U=\{U_X,U_C\}$.  
\end{assumption}

\subsubsection{Comparisons with CEVAE and its variants}
CEVAE and its variants treat $X$ as a proxy for unmeasured variables and propose a variational inference model to recover $\{ U_{IV},  U_X, U_A \}$ \citep{louizos2017cevae, zhang2021tedvae, wu2022betaintactvae}. Nevertheless, they ignored the unmeasured confounders $U_C$ independent of $X$, which led to biased treatment effect estimation. By empirical experiments, \citet{rissanen2021criticalcevae} claims that CEVAE seems to work reliably under simple scenarios and does not recover all latent variables from proxy variables $X$. Therefore, as illustrated in Figure \ref{figure2}(b), we propose a latent-variable module to recover the part of latent IVs $U_{IV}$ and latent confounders $U_X$ instead of full confounders or full IVs, and plug it into CB-IV, named CB-IV-L. As a distinction, we use the symbol $L=\{U_{IV}, U_X\}$ to denote the latent variables of interest, and we are not concerned with the recovery of $U_A$ and $U_C$. 

\subsubsection{CB-IV-L Algorithm}
\label{sec:algorithm2}
Like previous works on latent variables for treatment effect estimation, such as CEVAE \citep{louizos2017cevae} and TEDVAE \citep{zhang2021tedvae}, our CB-IV-L also builds upon VAEs \citep{kingma2014auto,rezende2014sto}. A significant advantage of VAEs is that they make substantially weaker assumptions about the data generating process and the structure of the latent confounders \citep{kingma2014auto, louizos2017cevae}. Next, we propose a latent-variable module to infer the complex nonlinear relationships between $\{L,X,T\}$ and approximately recover the conditional probability distribution $Pr(L|X,T)$ with $L = \{ U_{IV},  U_X, \}$, as shown in Figure~\ref{figure2}(b).

\textbf{Latent-Variable Module}.
To this end, our goal is to learn the posterior distribution $Pr(L|X,T)$ for the set of latent factors with $L = \{U_{IV}, U_X\}$ as illustrated in Figure~\ref{figure2}(b), where $U_{IV}$ and $U_X$ correspond latent IVs and latent confounders, respectively. Note that we do not impose the disentanglement assumption with perfectly separated vectors $U_{IV}$ and $U_X$, as the CB-IV algorithm can apply to mixed variables $L$ as shown in Figure~\ref{figure2}(a). Thus, we tend to reconstruct the latent variables $L = \{U_{IV}, U_X\}$ as mixed data for treatment effect estimation. 
Following standard VAE design, the prior distribution $Pr(L)$ is chosen as Gaussian distributions \citep{kingma2014auto}:
\begin{eqnarray}
\label{eq:gauss}
Pr(l_i) = \prod_{j=1}^{m_L} \mathcal{N}(l_{ij} | 0,1), l_i \in L,
\end{eqnarray}
where, $l_{ij}$ denotes the $j$-th coordinate in vector $l_i$ of unit $i$, and $m_L$ is the dimension of the latent variables, i.e., $L \subset \mathbb{R}^{m_L}$. 

\textbf{Decoder}. As shown in Figure~\ref{figure2}(b), the data generation process $\{X,T,Y\}$ follows $Pr(X|U_{IV},  U_X, U_A)$, $Pr(T|U_{IV},  U_X, U)$, and $Pr(Y|T, U_X, U_A, U)$. Our goal is to learn the posterior distribution $Pr(L|X,T)$ with $L=\{U_{IV},  U_X\}$. In addition, $U$'s information is difficult to recover perfectly, because the noise proxy variable $X$ cannot provide supervision information for it. Therefore, we use the symbol $E$ to represent the effects of $U$ and $U_A$ with distribution $\mathcal{N}(E | 0,1)$. For simplicity, we embed it into $L$, and marked it as $L^e = \{L, E\}$. Specifically, the generative models for $\{X,T,Y\}$, i.e., decoder, are described as:
\begin{eqnarray}
\label{eq:gauss2}
Pr(x_i|l^e_i)=\prod_{j=1}^{m_X}Pr(x_{ij}|l^e_i), Pr(t_i|l^e_i), Pr(y_i|t_i, l^e_i).
\end{eqnarray}
Because exact inference would be intractable, for each coordinate in $\{x_{i1},\cdots,x_{im_X},t_i,y_i\}$, we employ neural network-based variational inference to approximate it. 
Next, we take $x_{ij}$ as an example to build inference model $Pr(x_{ij}|l^e_i)$, and  
build $Pr(t_i|l^e_i)$, $Pr(y_i|t_i, l^e_i)$ similarity. 

If $x_{ij}$ is a discrete variable with $K$ values, we model $Pr(x_{ij}|l^e_i)$ with neural networks $f_{x_{ij}}(\cdot): \mathbb{R}^{m_L} \rightarrow (0,1)^K$:
\begin{eqnarray}
\label{eq:pr_d}
Pr(x_{ij}|l^e_i)=\text{D}(\sigma(f_{x_{ij}}(l^e_i))),
\end{eqnarray}
where $\sigma(\cdot)$ is the logistic function that map $f_{x_{ij}}(l^e_i)$ as the probability of each discrete value, $\text{D}$ denotes the discrete distribution. 
If $x_{ij}$ is a continuous variable, we model $Pr(x_{ij}|l^e_i)$ with neural networks $\mu_{X_{ij}}(\cdot), \sigma_{X_{ij}}(\cdot): \mathbb{R}^{m_L} \rightarrow \mathbb{R}$:
\begin{eqnarray}
\label{eq:pr_c}
Pr(x_{ij}|l^e_i)=\mathcal{N}(\mu_{X_{ij}}(l^e_i),\sigma_{X_{ij}}^2(l^e_i)),
\end{eqnarray}
where the mean $\mu_{X_{ij}}(l^e_i)$ and the standard deviation $\sigma_{X_{ij}}(l^e_i)$ are the parameters of the normal distribution $\mathcal{N}$. Then, we can model $Pr(t_i|l^e_i)$, $Pr(y_i|t_i, l^e_i)$ similarity. 

\textbf{Encoder}. By the definition of the model in Figure~\ref{figure2}(b), we can see that the true posterior over $L^e$ depends on $X,T,Y$. To emphasize that our goal is to learn latent IVs $U_{IV}$ and latent confounders $U_X$, we employ the following posterior approximation\footnote{\citet{louizos2017cevae} employs $Pr(L^e|X,T,Y)$ in CEVAE, \citet{zhang2021tedvae} employs $Pr(L^e|X)$ in TEDVAE, \citet{wu2022betaintactvae} employs $\mathbb{E}_{Pr(T,Y|X)}Pr(L^e|X,T,Y)$ in $\beta$-Intact-VAE. In this paper, we use $\{X, T\}$ to approximate $Pr(L^e|X,T)$, since we don't care about the recovery of $\{U_C,U_A\}$ from $Y$. } $Pr(L^e|X,T)$, which only depends on $X,T$.
Then, we implement variational inference along with inference networks $\mu_{L}(\cdot), \sigma_{L}(\cdot): \mathbb{R}^{m_X + 1} \rightarrow \mathbb{R}^{m_L}$:
\begin{eqnarray}
\label{eq:pr_z}
Pr(l_i|x_i, t_i)=\mathcal{N}(\mu_{x_{L}}(x_i, t_i),\sigma_{x_{L}}^2(x_i, t_i)),
\end{eqnarray}
where the mean $\mu_{x_{L}}(x_i, t_i)$ and the standard deviation $\sigma_{x_{L}}(x_i, t_i)$ are the parameters of the normal distribution $\mathcal{N}$.

\textbf{Objective}.
We can now form an evidence lower bound (ELBO) for the inference $\{Pr(x_i|l^e_i), Pr(t_i|l^e_i), Pr(y_i|t_i, l^e_i),Pr(l_i|x_i, t_i)\}$, the networks $\{\mu_{X_{ij}}(\cdot), \sigma_{X_{ij}}(\cdot),\mu_{T}(\cdot), \sigma_{T}(\cdot)$, $\mu_{Y}(\cdot)$, $\sigma_{Y}(\cdot),\mu_{L}(\cdot), \sigma_{L}(\cdot)\}$ can be optimized by maximizing the objective \citep{kingma2014auto,rezende2014sto}:
\begin{eqnarray}
\label{eq:elbo}
\mathcal{L}_{ELBO} 
&=& \sum_{i=1}^n \mathbb{E}_{Pr(l_i|x_i, t_i)}[\log Pr(x_i|l^e_i) + \log Pr(t_i|l^e_i)  + \log Pr(y_i|t_i, l^e_i)] \nonumber \\ 
&& - D_{KL}(Pr(l_i|x_i, t_i) \| Pr(l_i)), 
\end{eqnarray}
where we use KL divergence $D_{KL}$ as a measure of a difference between two probability distributions. By maximizing the bound, we approximately recover the conditional probability distribution $Pr(L^e|X,T)$ with $L^e = \{ U_{IV},  U_X, E\}$. The bound guarantees that latent IVs $U_{IV}$ and latent confounders $U_X$ are embedded into the latent representation $L^e$. Although $L^e$ may obtain some information from $\{U_C,U_A\}$, that does not lead to a biased CB-IV, because $\hat{T} \sim \hat{Pr}(T|L^e)$ and $\hat{T} \perp \{U_C, U_A\} | L^e$.

Therefore, we can use the latent-variable module to learn latent IVs $U_{IV}$ and latent confounders $U_X$ for CB-IV. 
Furthermore, we can then directly replace the treatment regression in stage 1 with the latent-variable module to obtain $\hat{T} \sim \hat{Pr}(T|L^e)$. In the outcome regression stage, we use the latent variables $L^e$ as input to the balance module and obtain balanced representations $C=f(L)$, as illustrated in Figure~\ref{figure2}(b). As long as we can accurately reconstruct the latent variables $L=\{U_{IV}, U_X\}$, then the treatment effect estimation from CB-IV would be asymptotically unbiased, we named this algorithm CB-IV-L.
One limitation is that CB-IV-L is reliable if and only if the latent-variable module performs well. CB-IV-L requires more training data, at least 5000, than CB-IV, to ensure the performance of the latent-variable module. 

Due to the small sample size ($n<5000$), CB-IV-L fails to reconstruct the latent variables for two semi-synthetic datasets IHDP \& Twins. Thus, we only report the network structure and parameters on the synthetic datasets Syn \& Demand and perform the corresponding experiments detailed in the next section. 
Tabel~\ref{tab:optimal} lists all network structures and hyper-parameters of the latent-variable module in CB-IV-L used for Syn \& Demand in the experiments section. 

\begin{table}[t]
\centering
\caption{Network Structures and Hyper-parameters of Latent-Variable Module.}
    \begin{tabular}{c | c c c}
        \hline
         {\textbf{Stage}} & {\textbf{Setting}} &  {\textbf{Syn}} &  {\textbf{Demand}} \\
        \hline
        \multirow{5}[8]{*}{\textbf{\shortstack{Latent-Variable \\ Module}}} 
          & $m_L$ & 5/10 & 5  \\
          & Epoch & 5/10 & 300  \\
          & Batchsize & 200 & 200 \\
          & MLPLayers & [64]*3 & [64]*3 \\
          & Activation & ELU & ELU \\
          & BatchNorm & False & False \\
          & Learning\_Rate & 0.0001 & 0.0001 \\
          & Optimizer & Adam & Adam \\
        \hline
    \end{tabular}\\
    \label{tab:optimal}%
\end{table}%

\section{Experiments}
In the conventional IV setting, we propose a Confounder Balanced IV regression (CB-IV) method to jointly remove the bias from both the unmeasured confounders with IV regression and the observed confounders by balancing.
To verify the advantages of the proposed algorithm, we evaluate CB-IV algorithm on both synthetic and real-world datasets, and demonstrate the effectiveness of our approach with both binary and continuous treatment settings. 
Nevertheless, due to the untestable exclusion assumption, no pre-defined IVs would be common cases in real-world scenarios, e.g., the mixed-variable and the latent-variable settings, detailed in Section~\ref{sec:extend}. 
To this end, we extend our CB-IV to the mixed-variable setting, and propose a latent variable module to develop a CB-IV-L to address the latent IVs problem. 
Based on the different assumptions for datasets, we divide the experiments into three parts to discuss the performance of the CB-IV algorithm in the conventional IV setting, the mixed variable setting, and the latent variable setting, respectively. 

\subsection{Baselines} \label{sec:baseline}
In the conventional IV setting and the mixed variable setting, we compare our CB-IV algorithm with two groups of methods. One group is \textbf{IV based methods}: (1) \emph{DeepIV-LOG} and \emph{DeepIV-GMM} \citep{1hartford2017deep}: In the first stage, DeepIV models the treatment network with logistic regression network (LOG) or gaussian mixture models (GMM);
(2) \emph{KernelIV} \citep{22singh2019kernel} and \emph{DualIV} \citep{23muandet2019dual}: 
they implement 2-stage regression with different dictionaries of basis functions from reproducing kernel Hibert spaces;
(3) \emph{OneSIV} \citep{27lin2019one}: OneSIV merges the two stages to leverage the outcome to estimate the treatment distribution; (4) \emph{DFIV} \citep{15xu2020dfiv}: DFIV uses neural networks to fit nonlinear models to replace the linear counterparts in the conventional 2SLS approach. 
The other group is \textbf{confounder balancing methods}: (1) \emph{DFL} \citep{15xu2020dfiv}:
DFL, an ablation experiment of DFIV, performs the nonlinear outcome regression directly without using instrumental variables; (2) \emph{Rep} and \emph{CFR} \citep{44johansson2016learning,45shalit2017estimating}: Both Rep and CFR learn the representation of the observed confounders, but the former does not make any constraints, and the latter requires the learned representation to be independent of the treatment; (3) \emph{DRCFR} \citep{49hassanpour2019learning}: DRCFR identifies and balances the confounders from all observed variables. \\
\textbf{Note that}: \emph{OneSIV} can be seen as an ablation version of our \emph{CB-IV} algorithm without confounder balancing, and \emph{Rep} and \emph{CFR} are the ablation versions of our \emph{CB-IV} algorithm without IV regression. 

Furthermore, unmeasured IVs are a more thorny problem in IV regression algorithms. In the latent variable setting, thus, we propose a variation inference module to recover latent IVs and latent variables $L=\{U_{IV}, U_X\}$. Then, we plug the latent variables $L$ into the above confounder balancing methods and CB-IV for estimating treatment effect and add the suffix '-L' to distinguish them from the original models, i.e., DFL-L, Rep-L, CFR-L, DRCFR-L, and CB-IV-L. Besides, because our model is built on CEVAE, we also compare our CB-IV-L algorithm with CEVAE-L. 

Next, we will prove the CB-IV approach's effectiveness in the conventional setting and then discuss the generalization performance of CB-IV in two more general settings. 

\subsection{Experiments on Conventional IV Setting} \label{sec:expiv}

\subsubsection{Experiments on Synthetic Datasets} 
\textbf{Dataset}. \label{synData}
In binary treatment cases, similar to \citep{49hassanpour2019learning}, we generate the synthetic datasets satisfying homogeneity assumption, as follows: the latent variables $\{Z,X,U\}$ drive from $Z_1,\cdots Z_{m_Z} \sim \mathcal{N}(0,\text{I}_{m_Z}), X_1,\cdots X_{m_X}, U_1,\cdots U_{m_U} \sim \mathcal{N}(0,\Sigma_{m_X+m_U})$ where $m_Z$, $m_X$ and $m_U$ are the dimensions of instruments $Z$, observed confounders $X$ and unobserved confounders $U$, respectively. $\text{I}_{m_Z}$ denotes ${m_Z}$ degree identity matrix,
$\Sigma_{m_X+m_U}=\text{I}_{m_X+m_U} * 0.95 + \mathds{1}_{m_X+m_U} * 0.05$ means that all elements except diagonal are 0.05 in the covariance matrix, and $\mathds{1}_{m_X+m_U}$ denotes ${m_X+m_U}$ degree all-ones matrix. 
The treatment variable $T$ and outcome variable $Y$ are generated as follows:
\begin{eqnarray}
\nonumber P(T \mid Z,X)  & = &   \frac{1}{1+\exp{\left(-(\sum_{i=1}^{m_Z}Z_iX_i+\sum_{i=1}^{m_X}X_i+\sum_{i=1}^{m_U}U_i)\right)}}, \\
\label{policy1} T  & \sim &  Bernoulli(P(T \mid Z,X)),m_X > m_Z\\
Y(T,X,U) & = & \frac{T}{m_X+m_U}(\sum_{i=1}^{m_X}X_i^2+\sum_{i=1}^{m_U}U_i^2) + \frac{1-T}{m_X+m_U}(\sum_{i=1}^{m_X}X_i+\sum_{i=1}^{m_U}U_i)
\label{policy2}
\end{eqnarray}
where $Bernoulli(P(T \mid Z,X))$ is the true logging policy of the treatment $T$. Eqs. (\ref{policy1}) \& (\ref{policy2}) is a common setting used by \citet{48hassanpour2019counterfactual, 49hassanpour2019learning, 50wu2022learning}.

As for continuous treatment cases, demand Datasets satisfying homogeneity assumption (that applied in DeepIV \citep{1hartford2017deep}, KernelIV \citep{22singh2019kernel}, DualIV \citep{23muandet2019dual} and DFIV \citep{15xu2020dfiv}) is a choice, and we report mean squared error (MSE) and its standard deviations over 10 trials: the outcome variabl is $Y=100+(10+T) X_1 \psi_{X_2}-2 T+E$; the treatment variable is $T=25+(Z+3) \psi_{X_2}+U$; $\psi_{X_2}=2\left((X_2-5)^{4} / 600+\exp \left[-4(X_2-5)^{2}\right]+X_2 / 10-2\right)$; where $X_1 \in\{1, \ldots, 7\}$, $X_2 \sim \operatorname{unif}(0,10)$, $Z, U \sim \mathrm{N}(0,1)$ and $E \sim \mathrm{N}\left(0.5 U, 0.75\right)$. In this case, the instrument variable is $Z$, the treatment variable is $T$, the observed variables are $\{X_1,X_2\}$, the outcome variable is $Y$, the unmeasured confounder is $\{U,E\}$.


\begin{table}[t]
\caption{The results of ATE estimation, including bias (mean(std)), in binary treatment cases on Synthetic data with different settings (Syn-$m_Z$-$m_X$-$m_U$).}
\vspace{-8pt}
\label{dimExp}
\begin{center}
\scalebox{0.85}{
\begin{tabular}{ccccc}
\hline
\multicolumn{5}{c}{\bf Within-Sample} \\
\hline
\bf Method & \bf Syn-1-4-4 & \bf Syn-2-4-4 & \bf Syn-2-10-4 & \bf Syn-2-4-10 \\
\hline
\bf DeepIV-LOG & 1.055(0.011) & 1.057(0.008) & 1.092(0.009) & 1.020(0.008) \\
\bf DeepIV-GMM & 0.934(0.011) & 0.874(0.019) & 0.768(0.023) & 0.925(0.017) \\
\bf KernelIV  & 0.495(0.056) & 0.457(0.054) & 0.765(0.028) & 0.624(0.062) \\
\bf DualIV & 1.469(0.072) & 1.423(0.076) & 1.719(0.076) & 1.534(0.073) \\
\bf OneSIV & 0.823(0.075) & 0.661(0.096) & 0.689(0.054) & 0.850(0.073) \\
\bf DFIV & 0.852(0.010) & 0.860(0.007) & 0.851(0.007) & 0.886(0.009) \\
\hline
\bf DFL & 0.840(0.002) & 0.851(0.002) & 0.838(0.002) & 0.831(0.004) \\
\bf Rep & 0.172(0.017) & 0.163(0.008) & 0.118(0.017) & 0.199(0.016) \\
\bf CFR & 0.172(0.016) & 0.158(0.015) & 0.105(0.020) & 0.198(0.018) \\
\bf DRCFR & 0.151(0.056) & 0.136(0.034) & \bf 0.063(0.044) & 0.154(0.032) \\
\hline
\bf CB-IV & \bf 0.038(0.071) & \bf 0.016(0.047) & 0.077(0.041) & \bf 0.009(0.065) \\
\hline
\hline
\multicolumn{5}{c}{\bf Out-of-Sample} \\
\hline
\bf Method & \bf Syn-1-4-4 & \bf Syn-2-4-4 & \bf Syn-2-10-4 & \bf Syn-2-4-10 \\
\hline
\bf DeepIV-LOG & 1.055(0.010) & 1.057(0.008) & 1.093(0.009) & 1.020(0.008) \\
\bf DeepIV-GMM & 0.933(0.011) & 0.874(0.019) & 0.768(0.023) & 0.925(0.017) \\
\bf KernelIV & 0.495(0.055) & 0.458(0.052) & 0.765(0.028) & 0.625(0.063) \\
\bf DualIV & 1.472(0.079) & 1.467(0.076) & 1.732(0.072) & 1.513(0.066) \\
\bf OneSIV & 0.822(0.076) & 0.661(0.095) & 0.690(0.053) & 0.851(0.073) \\
\bf DFIV & 0.851(0.009) & 0.860(0.007) & 0.851(0.007) & 0.886(0.009) \\
\hline
\bf DFL & 0.840(0.002) & 0.851(0.002) & 0.838(0.002) & 0.831(0.004) \\
\bf Rep & 0.172(0.016) & 0.164(0.009) & 0.116(0.015) & 0.199(0.014) \\
\bf CFR & 0.172(0.015) & 0.159(0.018) & 0.103(0.019) & 0.198(0.016) \\
\bf DRCFR & 0.151(0.055) & 0.137(0.035) & \bf 0.062(0.045) & 0.154(0.032) \\
\hline
\bf CB-IV & \bf 0.037(0.075) & \bf 0.017(0.046) & 0.075(0.040) & \bf 0.010(0.064) \\
\hline
\end{tabular}
}
\vspace{-8pt}
\end{center}
\end{table}


\noindent \textbf{Results in binary treatment cases}. 
The results of treatment effect estimation in binary treatment cases are reported in Table \ref{dimExp}, where we use Syn-$m_Z$-$m_X$-$m_U$ to denote the synthetic dataset with $m_Z$ instruments, $m_X$ observed confounders, and $m_U$ unobserved confounders. For each setting (such as \emph{Syn-1-4-4}, \emph{Syn-2-4-4}, \emph{Syn-2-10-4}, \emph{Syn-2-4-10}), we sample 10,000 units and perform 10 replications to report the mean and the standard deviation (std) of the bias of the average treatment effect (ATE) estimation, where \emph{within-sample} error is computed over the training sets, and \emph{out-of-sample} error is over the test set. From the results in Table \ref{dimExp}, we have the following observations: (1) For IV-based methods, more valid IVs would bring more accuracy on treatment effect estimation by comparing with the results of setting \emph{Syn-1-4-4} and setting \emph{Syn-2-4-4}. (2) For confounder balancing methods, a high dimension of unmeasured confounder would lead to poor performance by comparing with the results of setting \emph{Syn-2-4-4} and setting \emph{Syn-2-4-10}. (3) The existence of observed confounders would result in the poor performance of the IV-based methods, even worse than the confounder balancing-based methods because traditional IV-based methods ignored the bias of observed confounders in their second-stage regression. (4) Considering confounder balancing in IV regression, our CB-IV improves considerably over the traditional IV-based methods and achieves better performance than confounder balancing methods in most settings. When the observed confounders are high-dimensional, the low-dimensional instruments' information might get lost, and CB-IV would be equivalent to CFR. 


\begin{figure}[t]
\begin{center}
\includegraphics[width=0.7\linewidth]{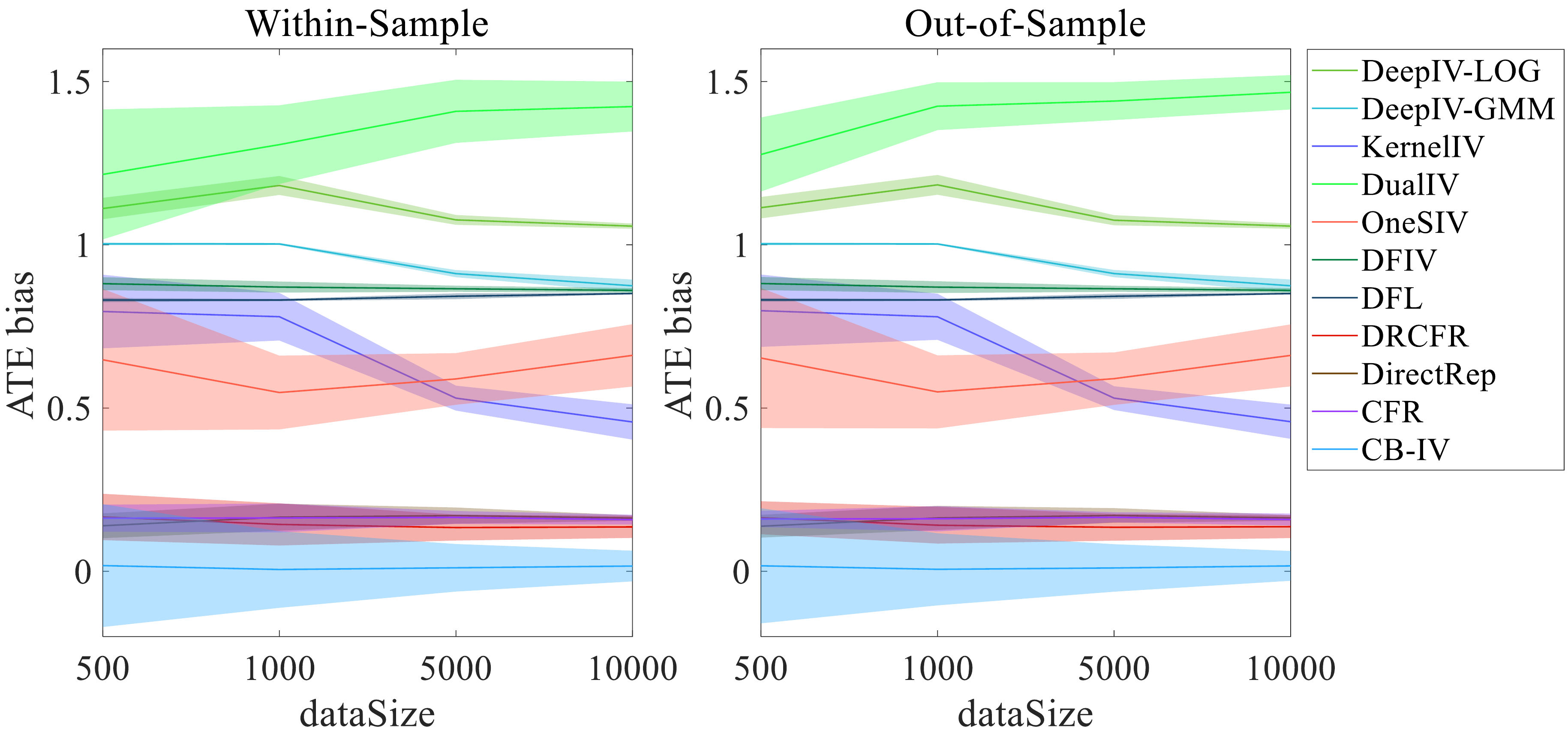}
\end{center}
\vskip -0.1in
\caption{Results of CB-IV on Syn-2-4-4 by varying sample size.}
\label{figure3}
\end{figure}


As a data-driven representation learning method, CB-IV requires more training data to ensure performance. Hence we implement experiments with different data sizes (500, 1000, 5000, 10000) on \emph{Syn-2-4-4} to study its impact on model performance. Figure (\ref{figure3}) shows that the bias of the average treatment effect estimation of CB-IV is low in different data sizes, but the variance is huge above small data sets. As the number of data increases, the variance of CB-IV will decrease linearly. When the amount of data exceeds 5000, the upper bound of CB-IV's estimation will be lower than the lower bound of all baselines. 
In conclusion, our method relies more on a large amount of data.


\begin{table}[t]
\caption{The results of potential outcome estimation, including MSE (mean(std)), in continuous treatment cases on Demand datasets with different settings (Demand-$\gamma$-$\lambda$).}
\vspace{-5pt}
\label{demand1}
\begin{center}
\scalebox{0.85}{
\begin{threeparttable}
\begin{tabular}{cccc}
\hline
\multicolumn{4}{c}{\bf Within-Sample} \\
\hline
\bf Method & \bf Demand-0-1 & \bf Demand-0-5 & \bf Demand-5-1 \\
\hline
\bf DeepIV-GMM & 1356(343.5) & 3102(744.4) & 1465(253.3) \\
\bf KernelIV & 1526(141.7) & $>$9999 & 1428(227.3) \\
\bf DualIV & $>$9999 & $>$9999 & $>$9999 \\
\bf OneSIV & $>$9999 & $>$9999 & $>$9999 \\
\bf DFIV & 195.2(9.342) & 1205(1740) & 197.2(16.80) \\
\hline
\bf DFL & 195.9(11.13) & 1159(1902) & 200.3(8.916) \\
\bf Rep & 191.2(5.514) & 888.6(1077) & 440.1(117.3) \\
\bf CFR & 193.3(5.561) & 465.3(181.4) & 449.6(161.0) \\
\bf DRCFR & 427.2(162.0) & 391.6(28.21) & 405.8(105.9) \\
\hline
\bf CB-IV & \bf 165.0(5.959) & \bf 234.1(30.06) & \bf 167.7(6.783) \\
\hline
\hline
\multicolumn{4}{c}{\bf Out-of-Sample} \\
\hline
\bf Method & \bf Demand-0-1 & \bf Demand-0-5 & \bf Demand-5-1 \\
\hline
\bf DeepIV-GMM & 1006(313.7) & 2829(724.6) & 1151(284.1) \\
\bf KernelIV & 994.9(146.2) & 5435(435.2) & 1004(216.7) \\
\bf DualIV & $>$9999 & $>$9999 & $>$9999 \\
\bf OneSIV & $>$9999 & $>$9999 & $>$9999 \\
\bf DFIV & 190.5(8.977) & 668.3(566.7) & 196.2(16.66) \\
\hline
\bf DFL & 182.9(11.52) & 597.6(622.1) & 189.7(7.422) \\
\bf Rep & 193.9(7.380) & 689.6(692.1) & 489.9(121.1) \\
\bf CFR & 192.0(8.932) & 417.3(123.5) & 469.7(140.7) \\
\bf DRCFR & 532.4(199.5) & 497.3(26.37) & 470.5(143.4) \\
\hline
\bf CB-IV & \bf 172.9(5.340) & \bf 224.3(18.06) & \bf 165.8(7.142) \\
\hline
\end{tabular}
\begin{tablenotes}    
\footnotesize               
    \item[*] We remove the baseline DeepIV-LOG for Demand dataset, since continuous treatment cannot be regressed with logistic functions. 
\end{tablenotes} 
\end{threeparttable}
}
\end{center}
\vspace{-15pt}
\end{table}



\begin{table}[t]
\caption{The results of latent outcome estimation, including MSE (mean(std)), in continuous treatment cases on Demand-0-1.}
\vspace{-10pt}
\label{continuous}
\begin{center}
\scalebox{0.85}{
\begin{threeparttable}
\begin{tabular}{ccccc}
\hline
\multicolumn{5}{c}{\bf Within-Sample} \\
\hline
\bf Method & \bf 500 & \bf 1000 & \bf 5000 & \bf 10000 \\
\hline
\bf DeepIV-GMM & $>$9999 & $>$9999 & 3163.3(266.4) & 1356(343.5) \\
\bf KernelIV & 3078(647.2) & 2363(270.7) & 1692(72.6) & 1526(141.7) \\
\bf DualIV & $>$9999 & $>$9999 & $>$9999 & $>$9999 \\
\bf OneSIV & $>$9999 & $>$9999 & $>$9999 & $>$9999 \\
\bf DFIV & 240.0(381.7) & 152.4(52.83) & 198.9(30.62) & 195.2(9.342) \\
\hline
\bf DFL & 141.4(26.42) & 173.2(29.90) & 196.8(17.82) & 195.9(11.13) \\
\bf Rep & 138.7(24.01) & 153.4(16.67) & 193.0(12.87) & 191.2(5.514) \\
\bf CFR & 126.9(20.98) & 161.7(20.99) & 191.6(10.24) & 193.3(5.561) \\
\bf DRCFR & 705.5(462.9) & 503.0(240.5) & 419.0(126.1) & 427.2(162.0) \\
\hline
\bf CB-IV & \bf 117.6(23.25) & \bf 142.0(16.11) & \bf 164.6(7.443) & \bf 165.0(5.959) \\
\hline
\hline
\multicolumn{5}{c}{\bf Out-of-Sample} \\
\hline
\bf Method & \bf 500 & \bf 1000 & \bf 5000 & \bf 10000 \\
\hline
\bf DeepIV-GMM &  $>$9999 &  $>$9999 & 3360(483.8) & 1006(313.7) \\
\bf KernelIV & 2859(660.9) & 2280(547.9) & 1142(170.3) & 994.9(146.2) \\
\bf DualIV &  $>$9999 &  $>$9999 &  $>$9999 &  $>$9999 \\
\bf OneSIV &  $>$9999 &  $>$9999 &  $>$9999 &  $>$9999 \\
\bf DFIV & 764.4(415.1) & 404.9(133.1) & 214.4(30.66) & 190.5(8.977) \\
\hline
\bf DFL & 358.7(47.32) & 261.3(35.68) & 192.7(14.46) & 182.9(11.52) \\
\bf Rep & 271.8(25.76) & \bf 222.3(9.575) & 199.8(5.453) & 193.9(7.380) \\
\bf CFR & \bf 266.2(28.45) & 225.9(11.75) & 195.8(11.338) & 192.0(8.932) \\
\bf DRCFR & 799.8(467.5) & 621.7(275.9) & 511.0(155.04) & 532.4(199.5) \\
\hline
\bf CB-IV & 291.4(39.33) & 229.1(42.22) & \bf 179.4(4.221) & \bf 172.9(5.340) \\
\hline
\end{tabular}
\begin{tablenotes}    
\footnotesize               
    \item[*] The results of IV-based methods are consistent with those of the report in DeepIV \citep{1hartford2017deep}, KernelIV \citep{22singh2019kernel}, DualIV \citep{23muandet2019dual} and DFIV \citep{15xu2020dfiv}. The difference is that they scale the results by log10, but we don't. 
\end{tablenotes} 
\end{threeparttable}
}
\end{center}
\vspace{-20pt}
\end{table}


\noindent \textbf{Results in continuous treatment cases}. \label{exp:continuous}
We adjust the difficulty of the simulation and perform experiments to increase the importance of instrumental variables in the structure function of $T$ (e.g., adjust $\gamma$ and $\lambda$ in $T=25+\gamma Z+(\lambda Z+3) \psi_{X_2}+U$), we name it as Demand-$\gamma$-$\lambda$. \emph{Demand-0-1} is the original Demand data with $T=25+(Z+3) \psi_{X_2}+U$. In \emph{Demand-0-5} with $T=25+(5 * Z+3) \psi_{X_2}+U$, we increase the information of the instrumental variable and amplify the confounding bias. As for \emph{Demand-5-1} with $T=25+5*Z + (Z+3) \psi_{X_2}+U$, we increase the information of the instrumental variable but keep the confounding bias unchanged.

The experimental results (reported in Table \ref{demand1}) show that (i) if the information of instrumental variables and confounders increases, all methods will become worse, but the confounder balance-based methods (e.g., CFR) still perform much better than the pure IV based methods (e.g., DeepIV). (ii) If we only increase the information of the instrumental variable, the results of the pure IV-based methods and our CB-IV are almost unchanged due to the same confounding bias. However, the balanced representation methods are basically worse, which is a magical phenomenon. One conjecture is that the fluctuation of $T$ affects the change of $Y$. Perhaps we should regularize the treatment variables and outcome variables before regressing them. Anyway, the confounding bias from the treatment regression stage is a critical problem in IV-based methods. 

Like the binary treatment studies in this paper, on this classical simulation data Demand-0-1 (Table \ref{continuous}) with different data sizes (500, 1000, 5000, 10000), the confounder balancing-based methods (without using IV) still perform much better than the pure IV-based methods. Considering confounder balancing in IV regression, our CB-IV method improves considerably over the traditional IV-based methods and achieves better performance than confounder balancing-based methods in most settings. Nevertheless, our method still relies on large samples.


\begin{table}[t]
\caption{The results of ATE estimation, including bias (mean(std)), on real-world data with different settings (Data-$m_Z$-$m_X$-$m_U$).}
\label{realExp}
\vskip 0.21in
\begin{center}
\scalebox{0.80}{
\begin{threeparttable}
\begin{tabular}{ccccc}
\hline
\multicolumn{5}{c}{\bf Within-Sample} \\
\hline
\bf Method & \bf IHDP-2-6-0 & \bf IHDP-2-4-2 & \bf Twins-5-8-0 & \bf Twins-5-5-3 \\
\hline
\bf DeepIV-LOG & 2.874(0.058) & 2.623(0.065) & 0.013(0.021) & 0.024(0.011) \\
\bf DeepIV-GMM & 3.776(0.032) & 3.740(0.040) & 0.019(0.005) & 0.022(0.004) \\
\bf KernelIV & 3.061(0.305) & 2.994(0.463) & - & -\\
\bf DualIV & 0.593(0.221) & 0.658(0.243) & - & -\\
\bf OneSIV & 1.725(0.375) & 1.741(0.342) & 0.008(0.019) & 0.008(0.017) \\
\bf DFIV & 3.554(0.089) & 3.622(0.104) & 0.027(0.001) & 0.026(0.000) \\
\hline
\bf DFL & 3.202(0.050) & 3.199(0.037) & 0.062(0.059) & 0.085(0.005) \\
\bf Rep & 0.068(0.056) & 0.460(0.071) & 0.017(0.017) & 0.019(0.025) \\
\bf CFR & 0.085(0.058) & 0.483(0.064) & 0.011(0.017) & 0.022(0.018) \\
\bf DRCFR & 0.055(0.064) & 0.434(0.069) & 0.011(0.022) & 0.012(0.017) \\
\hline
\bf CB-IV & \bf 0.012(0.388) & \bf 0.160(0.250) & \bf 0.007(0.027) & \bf 0.001(0.025) \\
\hline
\hline
\multicolumn{5}{c}{\bf Out-of-Sample} \\
\hline
\bf Method & \bf IHDP-2-6-0 & \bf IHDP-2-4-2 & \bf Twins-5-8-0 & \bf Twins-5-5-3 \\
\hline
\bf DeepIV-LOG & 2.876(0.055) & 2.623(0.069) & 0.014(0.021) & 0.024(0.011) \\
\bf DeepIV-GMM & 3.777(0.035) & 3.739(0.042) & 0.019(0.005) & 0.022(0.004) \\
\bf KernelIV & 3.070(0.306) & 3.023(0.440) & - & -\\
\bf DualIV & 0.564(0.266) & 0.715(0.355) & - & -\\
\bf OneSIV & 1.729(0.372) & 1.735(0.343) & 0.008(0.019) & 0.008(0.017) \\
\bf DFIV & 3.554(0.090) & 3.623(0.106) & 0.027(0.001) & 0.026(0.000) \\
\hline
\bf DFL & 3.204(0.050) & 3.199(0.038) & 0.062(0.058) & 0.085(0.005) \\
\bf Rep & 0.061(0.082) & 0.457(0.076) & 0.016(0.018) & 0.019(0.025) \\
\bf CFR & 0.079(0.081) & 0.480(0.069) & 0.011(0.016) & 0.022(0.018) \\
\bf DRCFR & 0.045(0.095) & 0.432(0.067) & 0.011(0.022) & 0.012(0.017) \\
\hline
\bf CB-IV & \bf 0.015(0.393) & \bf 0.158(0.254) & \bf 0.006(0.027) & \bf 0.002(0.025) \\
\hline
\end{tabular}
\begin{tablenotes}    
\footnotesize               
    \item[*] Most confounders are discrete variables and the outcome is binary variable in Twins data. The results of kernel-based IV methods in Twins are NaN. We use '-' to denote it. 
\end{tablenotes}     
\end{threeparttable}
}
\vskip -0.12in
\end{center}
\end{table}


\subsubsection{Experiments on Real-World Datasets}
\textbf{Dataset}.
We also check the performance of the CB-IV method with experiments on two real-world datasets, which are adopted in \citet{51yao2018representation, 50wu2022learning}: IHDP tends to evaluate the effect of a specialist home visit on premature infants' cognitive test scores, and Twins aims to estimate the effect of the weight in twins on the infant's mortality. 

The Infant Health and Development Program ({IHDP}\footnote{\url{http://www.fredjo.com/}}) comprises 747 units (139 treated, 608 control). To develop the instrument variables, we generate 2-dimension random variables for each unit, i.e., $Z_1,\cdots Z_{m_Z} \sim \mathcal{N}(0,I_{m_Z}), m_Z = 2$. Then, we select 6 variables from the original data as the confounders, including $m_X$ variables as observed 
confounders $X$ and $m_U$ as unobserved $U$, where $m_X+m_U=6$. The treatment assignment policy is $P(T \mid Z,X) = \frac{1}{1+\exp{\left(-(\sum_{i=1}^{m_Z}Z_iX_i+\sum_{i=1}^{m_X}X_i)+\sum_{i=1}^{m_U}U_i)\right)}}, T \sim Bernoulli(P(T \mid Z,X))$. 

{Twins}\footnote{\url{http://www.nber.org/data/}} dataset is derived from all twins born in the USA between the years 1989 and 1991 \cite{64almond2005costs}. Similar to \citet{51yao2018representation}, we select 5271 records from same-sex twins who weighed less than 2000 grams and had no missing characteristics. Then we generate 5-dimension random variables as the instrument variables and obtain $m_X$ variables as observed confounders $X$ and $m_U$ as unobserved $U$ to design the treatment $T$ according to the policy in Eq. (\ref{policy1}).  

\noindent \textbf{Results}.
We conduct our experiments over the 100 realizations of IHDP and 10 realizations of Twins with a 63/27/10 proportion of train/validation/test splits. In each realization, we shuffle the data and then redivide it into train/validation/test splits to simulate as many different data distributions as possible. 
\textbf{Data-$m_Z$-$m_X$-$m_U$} means that there are $m_Z$ dimension instruments, $m_X$ observed confounders and $m_U$ unobserved confounders in the corresponding Data.
We report the results in Table \ref{realExp}, including the mean and standard deviation (std) of the average treatment effect estimation bias. 

In the dataset without unmeasured confounders (\emph{IHDP-2-6-0} and \emph{Twins-5-8-0}), the performance of CB-IV is better than confounder balance methods (DRCFR, CFR),  better than two-head methods (Rep), and the IV methods (DeepIV, KernelIV, DFIV) are the worst. DualIV and OneSIV perform best in the traditional IV methods on \emph{IHDP} and \emph{Twins}, respectively. 
When there are unmeasured confounders (\emph{IHDP-2-4-2} and \emph{Twins-5-5-3}), it is evident that the performance of the confounder balance methods decreases a lot. Still, the performance of CB-IV and IV methods are almost unaffected, which is in line with our expectations.
CB-IV requires a larger amount of data to ensure variance convergence. Because the training set of IHDP has only 471 samples, CB-IV has a small bias but a large variance. Despite this, in the presence of unobserved confounders, the upper bound of the error of CB-IV is much lower than these baselines. In general, CB-IV achieves the best performance among all baselines.


\begin{table}[t]
\caption{The results of ATE estimation, including bias (mean(std)), in binary treatment cases on Synthetic data with different settings (Syn(vars used in stage 1)(vars used in stage 2)).}
\label{mode-syn}
\begin{center}
\scalebox{0.85}{
\begin{tabular}{ccccc}
\hline
\multicolumn{5}{c}{\bf Within-Sample} \\
\hline
\bf Method & \bf Syn(Z)(X) & \bf Syn(X)(X) & \bf Syn(Z,X)(Z,X) & \bf Syn(Z,X)(X) \\
\hline
\bf DeepIV-LOG & 1.055(0.006) & 1.054(0.007) & 1.059(0.009) & 1.057(0.008) \\
\bf DeepIV-GMM & 0.862(0.016) & 0.992(0.007) & 0.961(0.006) & 0.874(0.019) \\
{\bf KernelIV } & 0.964(0.070) & 0.865(0.174) & 0.890(0.157) & 0.457(0.054) \\
{\bf DualIV } & 0.658(0.561) & 1.611(0.495) & 1.763(0.042) & 1.423(0.076) \\
\bf OneSIV & 1.048(0.030) & 1.176(0.046) & 1.053(0.045) & 0.661(0.096) \\
\bf DFIV & 1.003(0.010) & 0.894(0.004) & 0.838(0.007) & 0.860(0.007) \\
\hline
\bf DFL & 0.843(0.002) & 0.843(0.002) & 0.842(0.002) & 0.843(0.002) \\
\bf Rep & 0.163(0.008) & 0.163(0.008) & 0.178(0.022) & 0.163(0.008) \\
\bf CFR & 0.158(0.015) & 0.158(0.015) & 0.177(0.023) & 0.158(0.015) \\
\bf DRCFR & \bf 0.136(0.034) & \bf 0.136(0.034) & 0.141(0.054) & 0.136(0.034) \\
\hline
\bf CB-IV & 0.495(0.263) & 0.529(0.100) & \bf 0.115(0.072) & \bf 0.016(0.047) \\
\hline
\hline
\multicolumn{5}{c}{\bf Out-of-Sample} \\
\hline
\bf Method & \bf Syn(Z)(X) & \bf Syn(X)(X) & \bf Syn(Z,X)(Z,X) & \bf Syn(Z,X)(X) \\
\hline
\bf DeepIV-LOG & 1.055(0.005) & 1.055(0.007) & 1.059(0.010) & 1.057(0.008) \\
\bf DeepIV-GMM & 0.862(0.016) & 0.992(0.007) & 0.961(0.006) & 0.874(0.019) \\
{ \bf KernelIV } & 0.963(0.070) & 0.865(0.177) & 0.916(0.157) & 0.458(0.052) \\
{\bf DualIV } & 0.800(0.307) & 1.606(0.501) & 1.760(0.037) & 1.467(0.053) \\
\bf OneSIV & 1.048(0.030) & 1.176(0.045) & 1.053(0.045) & 0.661(0.095) \\
\bf DFIV & 1.003(0.009) & 0.894(0.004) & 0.838(0.006) & 0.860(0.007) \\
\hline
\bf DFL & 0.843(0.002) & 0.843(0.002) & 0.842(0.002) & 0.843(0.002) \\
\bf Rep & 0.164(0.009) & 0.164(0.009) & 0.179(0.019) & 0.164(0.009) \\
\bf CFR & 0.159(0.018) & 0.159(0.018) & 0.178(0.023) & 0.159(0.018) \\
\bf DRCFR & \bf 0.137(0.035) & \bf 0.137(0.035) & 0.142(0.052) & 0.137(0.035) \\
\hline
\bf CB-IV & 0.493(0.261) & 0.528(0.099) & \bf 0.114(0.071) & \bf 0.017(0.046) \\
\hline
\end{tabular}
}
\end{center}
\end{table}


\begin{table}[t]
\caption{The results of ATE estimation, including bias (mean(std)), in Demand data with different settings (Demand(vars used in stage 1)(vars used in stage 2)).}
\label{mode-demand}
\begin{center}
\scalebox{0.85}{
\begin{tabular}{ccccc}
\hline
\multicolumn{5}{c}{\bf Within-Sample} \\
\hline
\bf Method & \bf Demand(Z)(X) & \bf Demand(X)(X) & \bf Demand(Z,X)(Z,X) & \bf Demand(Z,X)(X) \\
\hline
\bf DeepIV-GMM & 1405(515.5) & 3017(471.3) & 1699(824.7) & 1356(343.5) \\
\bf KernelIV   & $>$9999 & 7868(447.6) & 1236(188.0) & 1526(141.7) \\
\bf DualIV & $>$9999 & $>$9999 & $>$9999 & $>$9999 \\
\bf OneSIV & $>$9999 & $>$9999 & $>$9999 & $>$9999 \\
\bf DFIV       & 228.9(23.42) & 364.6(49.41) & 150.1(7.476) & 195.2(9.342) \\
\hline
\bf DFL & 195.9(11.13) & 195.9(11.13) & \bf 157.7(4.510) & 195.9(11.13) \\
\bf Rep & 191.2(5.514) & \bf 191.2(5.514) & 193.1(6.146) & 191.2(5.514) \\
\bf CFR & 193.3(5.561) & 193.3(5.561) & 188.2(5.110) & 193.3(5.561) \\
\bf DRCFR & 427.2(162.0) & 427.2(162.0) & 680.9(342.7) & 427.2(162.0) \\
\hline
\bf CB-IV & \bf 176.5(9.435) & 1591(39.24) & 174.3(8.301) & \bf 165.0(5.959) \\
\hline
\hline
\multicolumn{5}{c}{\bf Out-of-Sample} \\
\hline
\bf Method & \bf Demand(Z)(X) & \bf Demand(X)(X) & \bf Demand(Z,X)(Z,X) & \bf Demand(Z,X)(X) \\
\hline
\bf DeepIV-GMM & 947.5(317.5) & 1723(245.6) & 1505(701.5) & 1006(313.7)  \\
\bf KernelIV   & $>$9999 & 3886(245.4) & 1058(158.5) & 994.9(146.2) \\
\bf DualIV & $>$9999  & $>$9999 & $>$9999 & $>$9999 \\
\bf OneSIV & $>$9999  & $>$9999 & $>$9999 & $>$9999 \\
\bf DFIV       & 1009(132.8) & 1571.1(66.77) & 194.1(8.502) & 190.5(8.977) \\
\hline
\bf DFL & 182.9(11.52) & \bf 182.9(11.52) & \bf 175.9(7.033) & 182.9(11.52) \\
\bf Rep & 193.9(7.380) & 193.9(7.380) & 198.4(13.08) & 193.9(7.380) \\
\bf CFR & 192.0(8.932) & 192.0(8.932) & 195.6(13.05) & 192.0(8.932) \\
\bf DRCFR & 532.4(199.5) & 532.4(199.5) & 735.9(311.2) & 532.4(199.5) \\
\hline
\bf CB-IV & \bf 180.9(10.48) & 715.8(27.75) & 177.4(9.571) & \bf 172.9(5.340) \\
\hline
\end{tabular}
}
\end{center}
\end{table}


\subsection{Experiments on Mixed-Variable Setting} \label{sec:expmixed}

\textbf{Dataset}. In the real world, it is common for researchers to treat all observed variables as confounders due to the lack of prior knowledge of selecting IV. However, in practice, some non-confounding variables may be among the observed variables, such as instrumental variables, that affect only the treatment variables, and we call this case the mixed variable setting. 
To evaluate the effectiveness of our CB-IV algorithm in mixed data, containing at least one valid IV, we selected binary treatment case \emph{Syn-2-4-4} and continuous treatment case \emph{standard Demand-0-1} as the benchmarks from the above data described in Section~\ref{synData}. For noun simplification, in this subsection, we refer to them as Syn and Demand, respectively. 

In this section, we use \textbf{Data(vars used in stage 1)(vars used in stage 2)} to distinguish between different dataset settings, i.e., Data($Z$)($X$),Data($X$)($X$),Data($Z,X$)($Z,X$), and Data($Z,X$)($X$). 
Specifically, in the conventional IV setting with pre-defined IVs, CB-IV uses the concatenation of IVs and confounders $\{Z, X\}$ as input to the treatment regression stage and uses $X$  as input to confounder balancing in the outcome regression stage. We use Data($Z,X$)($X$) to represent this setting. In the mixed variable setting (Data($Z, X$)($Z,X$)), we can not distinguish IVs from observed covariates, and can only use mixed variables $\{Z,X\}$ as input for both the treatment and outcome regression stages. Data($X$)($X$) means that all observed data are confounders, i.e., no IVs presented in observed covariates. Besides, to evaluate the effect of interaction terms between IVs and covariates on treatment, we use only IVs $Z$ to regress treatment in Data($Z$)($X$). 

\noindent \textbf{Results}.
We sample 10000 units from \emph{Syn-2-4-4} and the standard \emph{Demand-0-1} to construct the datasets Data(vars used in stage 1)(vars used in stage 2) as well as perform 10 replications. We report the mean and the standard deviation on the bias of average treatment effect (ATE) estimation and average potential outcome estimation on different data settings in Table~\ref{mode-syn} \& \ref{mode-demand}. 

From the results, we have the following observations: (1) almost all IV methods achieve the best results on \emph{Syn(Z, X)(X)} \& \emph{Demand(Z, X)(X)}, compared with the other three settings, which is in line with our preliminary. That means higher-order interaction terms, between IVs and covariates in the treatment regression stage, can help us learn better and retain information about IVs; (2) in the mixed data \emph{Syn(Z,X)(Z,X)} \& \emph{Demand(Z,X)(Z,X)}, by balancing observed confounders in stage 2, our CB-IV algorithm outperforms all existing IV approaches and has comparable performance to the best confounder balancing methods (DRCFR in Syn; DFL in Demand). Through the ablation experiments (Rep and CFR), CB-IV reduces the error by 0.065 ($\downarrow 36.5\%$) in Syn and 18.2 ($\downarrow 9.3\%$) in Demand, thanks to the IV methods that eliminate unmeasured confounding; (3) in Syn(X)(X) and Demand(X)(X) settings, all instrumental variable methods including CB-IV fail to estimate unbiased causal effects, but confounder balancing methods still work. This means that when strong and valid IVs present in the data generation process are not observed, the treatment regression stage will introduce additional bias. How to recover the effects of strong latent IVs will be described in the next section; (4)
the results from Syn(Z)(X) and Demand(Z)(X) show the difference in the interaction terms for IV regression. If we only use instruments to regress $ZX$ term, it is easy to get a weak or even zero effect, i.e., ($\mathbb{E}[ZX|Z]=Z\mathbb{E}[X]=0$), and CB-IV would fail in Syn datasets.  In Demand datasets, CB-IV uses instruments to recover $\Scale[1.0]{\frac{25}{12}}Z$ to treatment effect and obtain outstanding performance.


\begin{table}[t]
\caption{The results of ATE estimation, including bias (mean(std)), in binary treatment cases on Synthetic data with different settings (Syn-$m_Z$-$m_X$-$m_U$).}
\label{dimExp1234}
\begin{center}
\scalebox{0.85}{
\begin{tabular}{ccccc}
\hline
\multicolumn{5}{c}{\bf Within-Sample} \\
\hline
\bf Method & \bf Syn-1-4-4 & \bf Syn-2-4-4 & \bf Syn-2-10-4 & \bf Syn-2-4-10 \\
\hline
\bf CEVAE-L & 0.522(0.004) & 0.528(0.004) & 0.132(0.009) & 0.508(0.004) \\
\bf DFL-L & 0.839(0.002) & 0.842(0.003) & 0.842(0.001) & 0.835(0.001) \\
\bf Rep-L & 0.077(0.016) & 0.086(0.027) & 0.034(0.021) & 0.146(0.020) \\
\bf CFR-L & 0.079(0.021) & 0.084(0.016) & 0.050(0.021) & 0.146(0.018) \\
\bf DRCFR-L & 0.203(0.068) & 0.198(0.085) & 0.389(0.119) & 0.171(0.037) \\
\hline
\bf CB-IV & 0.038(0.071) & 0.016(0.047) & 0.077(0.041) & \bf 0.009(0.065) \\
\bf CB-IV-L & \bf 0.002(0.023) & \bf 0.016(0.033) & \bf 0(0.027) & 0.101(0.018) \\
\hline
\hline
\multicolumn{5}{c}{\bf Out-of-Sample} \\
\hline
\bf Method & \bf Syn-1-4-4 & \bf Syn-2-4-4 & \bf Syn-2-10-4 & \bf Syn-2-4-10 \\
\hline
\bf CEVAE-L & 0.523(0.003) & 0.527(0.004) & 0.129(0.009) & 0.507(0.004) \\
\bf DFL-L & 0.839(0.002) & 0.842(0.002) & 0.842(0.002) & 0.835(0.001) \\
\bf Rep-L & 0.078(0.017) & 0.086(0.026) & 0.034(0.021) & 0.146(0.022) \\
\bf CFR-L & 0.079(0.021) & 0.084(0.015) & 0.051(0.023) & 0.146(0.018) \\
\bf DRCFR-L & 0.203(0.067) & 0.196(0.080) & 0.391(0.119) & 0.175(0.038)  \\
\hline
\bf CB-IV & 0.037(0.075) & 0.017(0.046) & 0.075(0.040) & \bf 0.010(0.064) \\
\bf CB-IV-L & \bf 0.002(0.023) & \bf 0.016(0.036) & \bf 0(0.027) & 0.101(0.019) \\
\hline
\end{tabular}
}
\end{center}
\end{table}


\subsection{Experiments on Latent-Variable Setting} \label{sec:explatent}

As the results of Syn(X)(X) and Demand(X)(X) shown in Table~\ref{mode-syn} \& \ref{mode-demand}, when strong and valid IVs present in the data generation process are not observed, referred as latent variable setting, all instrumental variable methods including CB-IV fail to estimate unbiased causal effects. To address the latent IVs problem, we propose a variation inference module to recover latent IVs and latent variables $L=\{U_{IV}, U_X\}$, and plug the latent variables $L$ into CB-IV for estimating treatment effect. In this section, we use the suffix '-L' to show that we use latent variables $L$ as input to regress the treatment effect, such as DFL-L, Rep-L, CFR-L, DRCFR-L, and CB-IV-L. Besides, CEVAE-L denotes a traditional variational inference model for causal effect estimation. 

\noindent \textbf{Dataset}. 
As a data-driven module, the reconstruction of latent variables depends on a large number of samples, at least 5000. Due to the small sample size ($n<5000$), the latent variable method fails to reconstruct the latent variables for two semi-synthetic datasets IHDP \& Twins. 
Thus, in this subsection, to evaluate the effectiveness of our CB-IV-L algorithm in latent data, we adopt the binary treatment cases \emph{Syn-$m_Z$-$m_X$-$m_U$} and continuous treatment cases \emph{Demand-$\gamma$-$\lambda$} as benchmarks, which are described in Section~\ref{synData}. 
We sample 10000 units from each dataset and perform 10 replications to report the mean and the standard deviation on the bias of average treatment effect (ATE) estimation and average potential outcome estimation on different data settings in Table~\ref{dimExp1234} \& \ref{demand1234}.


\begin{table}
\caption{The results of latent outcome estimation, including MSE (mean(std)), in continuous treatment cases on Demand datasets with different settings (Demand-$\gamma$-$\lambda$).}
\label{demand1234}
\begin{center}
\scalebox{0.85}{
\begin{tabular}{cccc}
\hline
\multicolumn{4}{c}{\bf Within-Sample} \\
\hline
\bf Method & \bf Demand-0-1 & \bf Demand-0-5 & \bf Demand-5-1 \\
\hline
\bf CEVAE-L & $>$9999 & $>$9999 & $>$9999 \\
\bf DFL-L & 8.340(3.683) & 18.35(4.680) & 8.692(2.895) \\
\bf Rep-L & 15.70(13.11) & 13.45(5.544) & 24.07(19.52) \\
\bf CFR-L &  10.65(4.264) & 12.42(3.092) & 15.39(9.152) \\
\bf DRCFR-L & 442.3(153.0) & 953.0(1129) & 926.4(573.0) \\
\hline
\bf CB-IV & { 165.0(5.959) } & { 234.1(30.06) } & { 167.7(6.783) } \\
\bf CB-IV-L & \bf 4.757(2.510) & \bf 6.455(3.614) & \bf 4.300(1.641) \\
\hline
\hline
\multicolumn{4}{c}{\bf Out-of-Sample} \\
\hline
\bf Method & \bf Demand-0-1 & \bf Demand-0-5 & \bf Demand-5-1 \\
\hline
\bf CEVAE-L & $>$9999 & $>$9999 & $>$9999 \\
\bf DFL-L & 10.335(2.883) & 21.67(6.012) & 10.97(3.532) \\
\bf Rep-L & 17.99(9.253) & 17.05(6.173) & 36.38(37.35) \\
\bf CFR-L & 13.58(6.076) & 16.059(5.010) & 24.44(33.76) \\
\bf DRCFR-L & 475.6(160.1) & 927.1(1003) & 896.1(418.9)  \\
\hline
\bf CB-IV & { 172.9(5.340) } & { 224.3(18.06) } & { 165.8(7.142) } \\
\bf CB-IV-L & \bf 5.228(3.090) & \bf 7.975(3.887) & \bf 4.564(2.324) \\
\hline
\end{tabular}
}
\end{center}
\end{table}

\noindent \textbf{Results}.
The experiments in Table~\ref{dimExp1234} \& \ref{demand1234} are a supplement, without pre-defined IVs, to the CB-IV experiments in Section~\ref{synData}. 

Comparing with the results of binary treatment cases Syn-$m_Z$-$m_X$-$m_U$ in Table~\ref{dimExp} and \ref{dimExp1234}, confounder balancing methods with latent variables (Rep-L \& CFR-L) report significantly lower ATE bias than the original version (Rep-L, CFR-L), suggesting that the latent-variable model is able to recover some latent confounders $U_X$ from the noisy proxies $X$ to achieve better confounder balancing. 
An anomaly is that as the effect from IVs increases (by adding one unit of IV information), the original method improves on Syn-2-4-4 compared to Syn-1-4-4, but we find that the performance of the latent variable version of these models decreases on Syn-2-4-4 compared to Syn-1-4-4. This difference implies that our latent-variable model learns more information about non-confounders, i.e., latent IVs. To verify this, by introducing instrumental variable regression, we find that CB-IV-L with latent IVs further reduces the estimation error and even surpasses the CB-IV algorithm's performance with pre-defined IVs. By learning 
latent confounders, we reduce the bias from $X$-related confounders, and by learning latent IVs, we reduce the bias from $X$-independent unmeasured confounders. Although the CB-IV-L method does not have pre-defined IVs, it has a better model structure under the Assumption~\ref{ass:latent}. 
The results of experiments Syn-2-10-4 and Syn-2-4-10 show that more $X$ will help us learn the latent variables more effectively, while a large number of unmeasured confounders will make it more difficult for the model to recover all the latent variables, but still help us improve the effect from Rep-L and CFR-L. 

Note that CEVAE-L, DFL-L, and DRCFR-L fail to obtain an unbiased treatment effect estimation from the high-level and more complex latent variables. As \citet{rissanen2021criticalcevae} said, these methods seem to work reliably under simple scenarios, and do not work under complex scenarios. 
Comparing the results of continuous treatment cases Demand-$\gamma$-$\lambda$ in Table~\ref{demand1} and \ref{demand1234}, we can find that the above conclusions about the binary scenario also apply to the continuous scenario. By learning latent IVs and latent confounders, as shown in Figure~\ref{figure2}(b), CB-IV-L jointly remove the bias from both the unmeasured confounders $U_C$ with IV regression and the confounders $U_X$ by balancing. 

\section{Conclusion}

The majority of instrumental variable methods ignore the confounding bias in the outcome regression stage in nonlinear scenarios. A promising direction is to implement confounder balancing. Under sufficient identification assumption, we propose a Confounder Balanced IV Regression (CB-IV) algorithm to confirm this and solve two inverse problems under different Homogeneity Assumptions. 
Nevertheless, due to the untestable exclusion assumption, no pre-defined IVs would be common in real-world scenarios. To solve the problem of no pre-defined IVs from expert knowledge, we extend CB-IV to more general settings and develop a CB-IV-L algorithm by a latent-variable module under mixed-variable and latent-variable challenges. Extensive experiments show that the proposed method achieves state-of-the-art performance in the average treatment effect estimation.

\newpage

\appendix
\section{Theorem and Proof}
\label{app:theorem}

In this appendix we prove the theorem~\ref{theory_1} from
Section~\ref{sec:inverse}:

\noindent
{\bf Theorem} {\it 
(\textbf{Inverse Relationship of Eq. (\ref{complicated_function2})}). If the learned representation of observed confounders $C=f_\theta(X)$ is independent with the estimated treatment $\hat{T}$, then the counterfactual prediction function ${ h  }(T,C)$ can be identified with instrumental variables $Z$ and representation $C$:
\begin{eqnarray}
    { h  }(T,C) = g_1^C(T,C)+g_2(T)\mathbb{E}[g_3(U)|C]+\mathbb{E}[g_4(X,U)|C].
\end{eqnarray}
Then, we can establish an inverse relationship for ${ h  }(T,C)$ given $\mathbb{E}[Y \mid Z,C,X]$ and $P(T \mid Z,X)$, as follow:
\begin{eqnarray}
\mathbb{E}[Y \mid Z,C,X] = \int \left[ { h  }(T,C) \right] dP(T \mid Z,X),
\end{eqnarray}
where, $dP(T \mid Z,X)$ is the conditional treatment distribution.
} \hfill\BlackBox

\noindent
{\bf Proof}. 
In this paper, we model the causal relationship more general and relax the additive separability assumption to the multiplicative assumption (Eq. (\ref{complicated_function1})(\ref{complicated_function2})): 
\begin{eqnarray*}
    T&=& f_1(Z,X)+f_2(X,U) \\ 
    Y&=& g_1(T,X)+g_2(T)g_3(U)+g_4(X,U), Z \perp U,X
\end{eqnarray*}
Then, we expect use the disentangled representation $C \perp T, C = f_\theta(X)$ to approximate the structural equation $Y = g_1^C(T,C)+g_2(T)g_3(U)+g_4(X,U)$.

Treatment Regression Stage, we perform nonlinear regression from $\{Z,X\}$ to $T$ using deep neural networks:
\begin{eqnarray*}
    \mathbb{E}[T|Z,X] & = & \mathbb{E}[f_1(Z,X)+f_2(X,U)|Z,X] \\
    & = & \mathbb{E}[f_1(Z,X)|Z,X]+\mathbb{E}[f_2(X,U)|Z,X] \\
    & = & f_1(Z,X)+\mathbb{E}[f_2(X,U)|X]
\end{eqnarray*}
where $\mathbb{E}[f_1(Z,X)|Z,X]=f_1(Z,X)$, because $Z$ and $X$ are independent. We define the conditional treatment distribution as $ \hat{T} \sim P(T|Z,X)$.

Outcome Regression Stage, we perform linear/nonlinear regression from $\{Z,X\}$ to $Y = g_1^C(T,C)+g_2(T)g_3(U)+g_4(X,U)$ using deep neural networks:

\begin{eqnarray}
\hspace{-18pt}
    \label{eq:2003}
    \nonumber \mathbb{E}[Y|Z,C,X] & = & \mathbb{E}[g_1(T,X)+g_2(T)g_3(U)+g_4(X,U)|Z,C,X] \\
    \nonumber & { =  } & \mathbb{E}[g_1^C(T,f_\theta(X))+g_2(T)g_3(U)+g_4(X,U)|Z,C,X] \\
    \nonumber & = & \mathbb{E}[g_1^C(T,C)+g_2(T)g_3(U)+g_4(X,U)|Z,C,X] \\
    \nonumber & = & \mathbb{E}[g_1^C(T,C)|Z,X]+\mathbb{E}[g_2(T)g_3(U)|Z,C,X] + \mathbb{E}[g_4(X,U)|C/X] \\
    \nonumber & = & \int g_1^C(T,C)dP(T|Z,X)+\mathbb{E}[g_2(T)g_3(U)|Z,C,X] + \mathbb{E}[g_4(X,U)|C/X] \\
    & = & \int \left[ g_1^C(T,C) + \mathbb{E}[g_4(X,U)|C] \right] dP(T|Z,X)+\mathbb{E}[g_2(T)g_3(U)|Z,C,X] ,
\end{eqnarray}
where $P(T|Z,X)$ is the conditional treatment distribution, $\mathbb{E}[g_4^C(C,U)|X]$ is a constant for the specified $X$/$C$. 

Because $\mathbb{E}[g_1^C(T,C)|Z,X] = \int g_1^C(T,C)dP(T|Z,X)$: the completeness of $Pr({T \mid Z,X})$ and $Pr({Y \mid T,X})$ would guarantees uniqueness of the solution \citep{29newey2003instrumental}. The relationship in Equation (\ref{eq:2003}) defines an inverse
problem for $g_1$ in terms of two directly observable functions: $\mathbb{E}[Y|Z,X]$ and $P(T|Z,X)$. Eq. (5) in \citet{1hartford2017deep} and Eq. (6) in \citet{27lin2019one} use same relationship to solve the inverse problem: 
\begin{eqnarray*}
& h(T, X) \equiv g(T, X)+\mathbb{E}[e \mid X] \\
& \mathbb{E}[Y \mid X, Z] =\mathbb{E}[g(T, X) \mid X, Z]+\mathbb{E}[e \mid X] =\int \hbar(T, X) d F(T \mid X, Z)
\end{eqnarray*}
where, again, $d F(T \mid X, Z)$ is the conditional treatment distribution in \citet{1hartford2017deep,27lin2019one}.

If $C \perp g_2(\hat{T})$, then $g_2(\hat{T}) \perp g_3(U) \mid Z,C$:
\begin{eqnarray*}
   & & \mathbb{E}[\mathbb{E}[g_2(\hat{T})g_3(U) \mid Z,C] \mid Z,X]  \\
   &=& \mathbb{E}[ \mathbb{E}[g_2(\hat{T}) \mid Z,C] \mathbb{E}[g_3(U) \mid Z,C]  \mid Z,X] \\
   &=& \mathbb{E}[ g_2(\hat{T}) \mathbb{E}[g_3(U) \mid C] \mid Z,X] \\
   &=& \int g_2(T) \mathbb{E}[g_3(U) \mid C] dP(T|Z,X)
\end{eqnarray*}

Summarily, 
\begin{eqnarray*}
&& \mathbb{E}[Y|Z,C,X] = \mathbb{E}[h(T,C)|Z,C,X] \\ 
& = &\int \left[ g_1^C(T,C) +g_2(T)\mathbb{E}[g_3(U)|C] + \mathbb{E}[g_4(X,U)|X] \right] dP(T|Z,X)
\end{eqnarray*}

The counterfactual prediction function is ${ h  }(T,C) = g_1^C(T,X)+g_2(T)\mathbb{E}[g_3(U)|C]+\mathbb{E}[g_4(X,U)|X]$, and can be identified by IVs and balanced representation. 
Then, we can establish an inverse relationship for ${ h  }(T,C)$ given $\mathbb{E}[Y \mid Z,C,X]$ and $P(T \mid Z,X)$, as follow:
\begin{eqnarray}
\nonumber \mathbb{E}[Y \mid Z,C,X] = \int \left[ { h  }(T,C) \right] dP(T \mid Z,X) ,
\end{eqnarray}
where, $dP(T \mid Z,X)$ is the conditional treatment distribution.
\hfill\BlackBox

\vskip 0.2in
\bibliography{reference}

\end{document}